\pdfoutput=1
\documentclass[11pt, a4paper]{stanford}

\usepackage[authoryear, sort&compress, round]{natbib}
\usepackage{microtype}
\usepackage{hyperref}
\usepackage{url}
\usepackage{lipsum}

\definecolor{accentcolor}{HTML}{175E54}
\hypersetup{
	colorlinks=true,
	citecolor=accentcolor,
	linkcolor=accentcolor,
	urlcolor=accentcolor
}

\usepackage{xcolor}
\usepackage{graphicx}
\usepackage{float}
\usepackage[hypcap=false]{subcaption}
\usepackage{booktabs}
\usepackage{colortbl}
\usepackage{caption}
\usepackage{makecell}
\usepackage{xspace}

\usepackage{mdframed}
\usepackage{listings}
\usepackage{amsmath}
\usepackage{amssymb}
\usepackage{pythonhighlight}
\usepackage{svg}
\usepackage{enumitem}
\usepackage{wrapfig}

\definecolor{vscbg}{RGB}{30, 30, 30}           % Background
\definecolor{vsctext}{RGB}{212, 212, 212}      % Default text
\definecolor{vsckeyword}{RGB}{86, 156, 214}    % Keywords (while, if, else)
\definecolor{vscstring}{RGB}{206, 145, 120}    % Strings
\definecolor{vsccomment}{RGB}{106, 153, 85}    % Comments
\definecolor{vscfunction}{RGB}{220, 220, 170}  % Function names
\definecolor{vscnumber}{RGB}{181, 206, 168}    % Numbers
\definecolor{vscpunctuation}{RGB}{212, 212, 212} % Punctuation
\definecolor{vsclinenumber}{RGB}{133, 133, 133} % Line numbers

\definecolor{promptbg}{RGB}{248, 249, 250}
\definecolor{promptborder}{RGB}{209, 213, 219}
\definecolor{guidancebg}{RGB}{240, 248, 255}

\newmdenv[
backgroundcolor=promptbg,
linecolor=promptborder,
linewidth=1pt,
roundcorner=5pt,
innertopmargin=10pt,
innerbottommargin=10pt,
innerleftmargin=15pt,
innerrightmargin=15pt
]{promptbox}

\newmdenv[
backgroundcolor=guidancebg,
linecolor=promptborder,
linewidth=1pt,
roundcorner=3pt,
innertopmargin=8pt,
innerbottommargin=8pt,
innerleftmargin=12pt,
innerrightmargin=12pt
]{guidancebox}

\usepackage[textwidth=2cm, textsize=small]{todonotes}
\usepackage{xcolor}
\usepackage{graphicx}

\definecolor{hao}{RGB}{255,100,100} % Light red

\usepackage{amsthm}
\theoremstyle{definition}

\usepackage{afterpage}
\usepackage{fontawesome}

\title{Real-Time Reasoning Agents in Evolving Environments}

\author{Yule Wen$^{1*}$ \quad Yixin Ye$^{2*}$ \quad Yanzhe Zhang$^{3}$ \quad Diyi Yang$^{4}$ \quad Hao Zhu$^{4}$ \\ 
$^{1}$Tsinghua University \quad $^{2}$Shanghai Jiao Tong University \quad $^{3}$Georgia Institute of Technology \quad $^{4}$Stanford University \\
$^{*}$Co-leading authors, work done during internship at Stanford University.\\
\faMousePointer\quad Website:~\url{https://realtimegym.saltlab.stanford.edu}\\
\faGithub\quad Code:~\url{https://github.com/SALT-NLP/RealtimeGym}\\
\faDatabase\quad Data:~\url{https://huggingface.co/datasets/SALT-NLP/RealtimeGym}
}

\definecolor{darkred}{rgb}{0.5, 0, 0}
\definecolor{darkblue}{rgb}{0, 0, 0.5}
\definecolor{fast}{HTML}{2993F0} % 蓝色
\definecolor{slow}{HTML}{FFBD17} % 黄色
\definecolor{parallel}{HTML}{45B43B} % 绿色

\newcommand{\ourProblem}{real-time reasoning}
\newcommand{\ourBenchmark}{Real-Time Reasoning Gym\xspace}
\newcommand{\ourSystem}{AgileThinker\xspace}
    \begin{abstract}
        Agents in the real world must make not only logical but also \emph{timely} judgments. This requires continuous awareness of the dynamic environment: hazards emerge, opportunities arise, and other agents act, while the agent's reasoning is still unfolding. Despite advances in language model reasoning, existing approaches fail to account for this dynamic nature. We introduce \emph{real-time reasoning} as a new problem formulation for agents in evolving environments and build \textbf{\ourBenchmark} to demonstrate it. We study two paradigms for deploying language models in agents:
        (1) reactive agents, which employ language models with \emph{bounded reasoning computation for rapid responses}, and
        (2) planning agents, which allow \emph{extended reasoning computation for complex problems}.
        Our experiments show that even state-of-the-art models struggle with making logical and timely judgments in either paradigm.
        To address this limitation, we propose \textbf{\ourSystem}, which simultaneously engages \emph{both reasoning paradigms}.
        \ourSystem{} consistently outperforms agents engaging only one reasoning paradigm as the task difficulty and time pressure rise, effectively balancing reasoning depth and response latency.
        Our work establishes real-time reasoning as a critical testbed for developing practical agents and provides a foundation for  research in temporally constrained AI systems, highlighting a path toward real-time capable agents.
    \end{abstract}

\begin{document}
    \maketitle
    %\tableofcontents
    
    \begin{center}
                \includegraphics[width=0.83\textwidth]{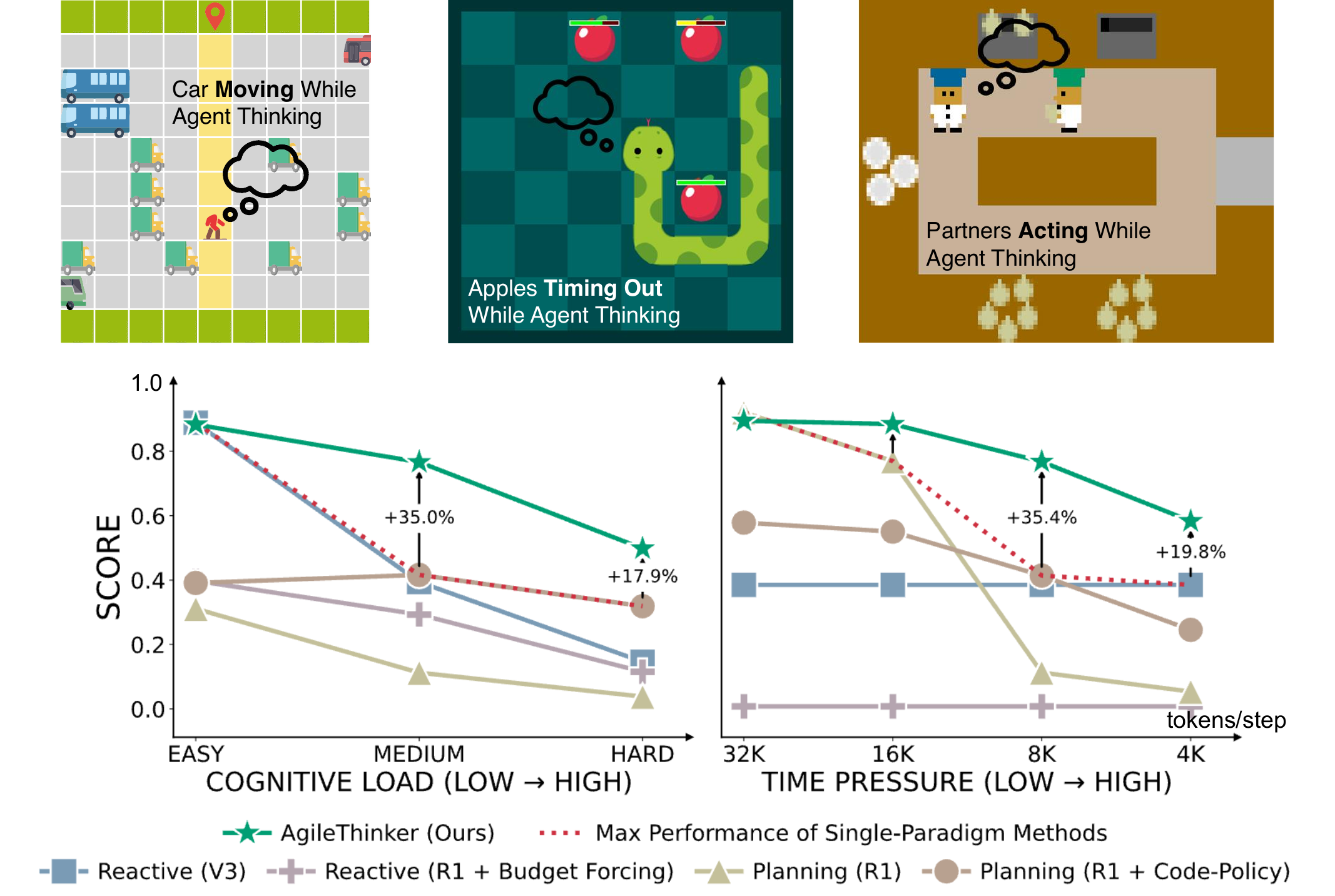}
        \captionof{figure}{Upper: Three real-time games, \textit{Freeway}, \textit{Snake}, and \textit{Overcooked}. Lower: Under \emph{cognitive load} and \emph{time pressure}, \ourSystem, engaging both \emph{reactive} and \emph{planning} reasoning, consistently outperforms agents that engage either of them. Scores are averaged across different games.}
        \label{fig:teaser}
    \end{center}
    % \afterpage{\thispagestyle{plain}}

    \section{Introduction}

\label{sec:introduction}

% 1. Problem Statement

Remember your first highway drive? White-knuckling the wheel, fixated on the car
ahead. Then suddenly your exit appears and you're three lanes over. This captures
how our brains work under pressure: reacting intuitively keeps us safe moment-to-moment,
but deliberate planning gets us where we need to go. This example highlights a fundamental
challenge: real-time reasoning \citep{Stanovich_West_2000, kahneman2011thinking, doi:10.1177/1745691612460685}.
Under time pressures, agents must simultaneously support timely reactions
and cognitively demanding long-term planning. Remarkably, humans excel at this
balancing act: shifting seamlessly between instinct and analysis as situations
demand. Expert drivers eventually handle both tasks effortlessly.

However, current Large Language Model (LLM)-based agents fail to live up to this challenge. 
Most existing work assume that the environments only change when the agents issue an action, ignoring the \emph{dynamic} nature of the world, which evolves in parallel to the agent's computation. 
As a result, despite great effort in improving agent planning with LLM reasoning, including
\citet{yao2022react,gou2023tora,putta2024agent,ferrag2025llm}, under this assumption, how to evaluate and improve the capability to make timely decisions is still an open question. 

To bridge this gap, we introduce \textbf{\ourBenchmark}, the first environment for
language agents to reason in dynamic environments (\S\ref{sec:real-time-reasoning-problem}). Our gym consists of three real-time
games: \emph{Freeway}, inspired by \href{https://en.wikipedia.org/wiki/Freeway_(video_game)}{the Atari game} under the same name,  \emph{Snake}, an adaptation of \href{https://en.wikipedia.org/wiki/Snake_(video_game_genre)}{a popular game}, and \emph{Overcooked}, a two-player version of the \href{https://en.wikipedia.org/wiki/Overcooked}{collaborative video game}. In each game, the state updates at a fixed rate regardless of whether the agent finishes its reasoning, and if no action is produced by the agent, a default action is used, simulating reasoning and acting in a real-time world. They challenge agents with different aspects of a dynamic environment: Freeway features dynamic
hazards with moving cars, Snake involves dynamic opportunities as food appears
and disappears, and Overcooked requires coordination with dynamic
partners who act on their own. \ourBenchmark is useful for studying different agent designs for real-time tasks.

% before the results, can we mention why this setting is beneficial and how it can help develop more realistic functioning real-world reasoners?
% 1. Easy to understand, hard to score high (extensive planning required)
% 2. 3 envs caputre the attributes of real-world real-time tasks; A demo test for realistic functioning real-world reasoners
% 3. Fine-grained test for different difficulty & time pressure
% 4. Fair test with \emph{elapsed token count}, independent from hardware conditions and focused on LLM systems themselves.

To compare different design choices of real-time reasoning agents, we study two paradigms: reactive agents and planning agents (\S\ref{sec:real-time-reasoning-agents}). Reactive agents ensure responsiveness by limiting computation, while planning agents are allowed to perform more extensive thinking. However, neither of them is perfect: planning agents cannot easily react to changes in the environment, and reactive agents fail to make strategic decisions.
We propose \textbf{\ourSystem} (\S\ref{sec:real-time-reasoning-agents}), a simple yet effective method that combines the strengths of \emph{both paradigms}. Unlike agents with one paradigm that must choose between speed and accuracy, \ourSystem\space runs two LLMs in two
parallel threads: a \emph{planning thread} performs extended reasoning over frozen game
states, and a \emph{reactive thread} outputs timely decisions within environmental
update time. Specifically, the reactive thread can reference partial reasoning traces from the ongoing planning process, enabling informed real-time decisions without waiting for complete analysis. This also differs from prior dual-system methods  \citep{zhang2025leveragingdualprocesstheory,liu2024llmpoweredhierarchicallanguageagent,christakopoulou2024agentsthinkingfastslow}, where either two systems operate independently, or one must wait for another to complete before accessing its outputs. 

In this paper, we study the following research questions: 
\begin{description}[leftmargin=2.5em, labelwidth=2em, labelsep=0.5em, itemindent=0pt]
    \item[RQ1] How do environment factors affect performance of agents in \ourBenchmark?
    %does \ourSystem outperform reactive agents under cognitive load, and planning agents under time pressure, and both agents under both conditions? \yxye{w. fixed model}
    \item[RQ2] How to balance reaction and planning resources in \ourSystem?
    \item[RQ3] How well do the results we get with simulation in \ourBenchmark match real-world walltime experiments? 
\end{description}

To study these questions, we manipulate the cognitive load and time pressure of the games, facilitating systematic evaluation across both dimensions. 
We evaluate different design choices for reactive and planning agents, including budget forcing \citep{muennighoff2025s1simpletesttimescaling} for reactive agents, code-as-a-policy \citep{codeaspolicies2022} for planning agents. 
For fair comparison across agent designs, we use one model family, focusing on DeepSeek V3 and R1 because they are open-source and provide transparent reasoning trajectories required by \ourSystem. 
To assess generality, we also experiment with proprietary models (App. \S~\ref{sec:gemini}) and observe similar performance trend.
Our results demonstrate that \ourSystem consistently outperforms single-paradigm methods by effectively balancing reactive and planning processes.%These findings translate to real-world scenarios, as confirmed by wallclock validation experiments.
This advantage is confirmed to translate to real-world scenarios through wall-clock time experiments. 
Ultimately, our work establishes a foundation for developing language agents capable of complex reasoning under time constraints, bringing AI systems closer to human-level performance in dynamic environments.

    \section{Real-Time Reasoning Gym}
\label{sec:real-time-reasoning-problem}

To evaluate the real-time reasoning capabilities of agents, we need an environment that
is (1) dynamic: the environment state continuously updates without waiting for the agent's decision; (2) cognitively challenging: the tasks should be challenging enough so that logical reasoning is needed; (3) reproducible: simulated environments to avoid non-negligible noise from hardware factors. 

\begin{figure}[!h]
    \centering
    \includegraphics[width=\linewidth]{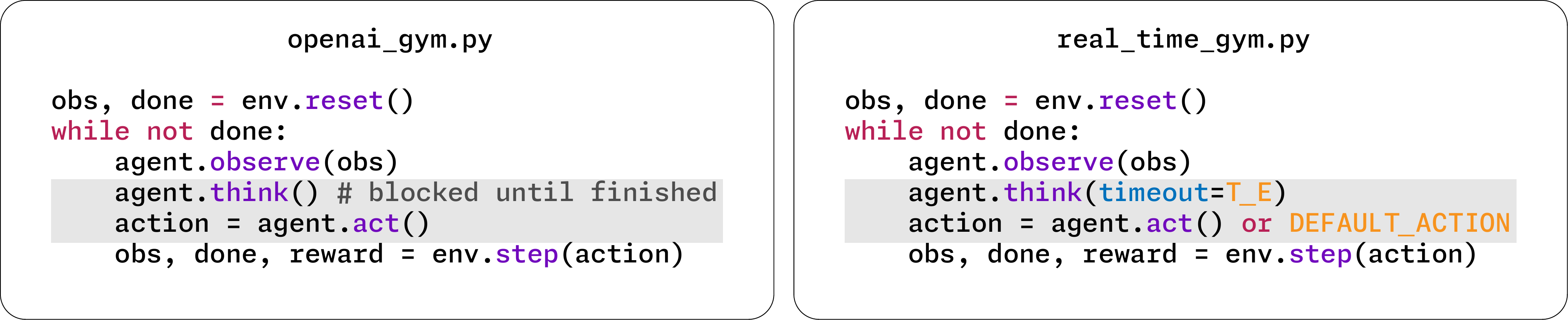}
    \caption{Agent loops in OpenAI Gym \citep{openaigym} and \ourBenchmark. Constants \texttt{T\_E} and \texttt{DEFAULT\_ACTION} will be explained in the following `time pressure' paragraph.}
    \label{fig:formulation}
\end{figure}

To achieve these three desiderata, we consider a new formulation of decision-making problem. Unlike conventional turn-based environments, where the environment steps only after the agent finishes thinking and produces an action (Figure \ref{fig:formulation} left), in \ourBenchmark, the environment steps forward at a fixed rate, even when the agent has not finished thinking. If no action is produced in time, a default action is applied (Figure \ref{fig:formulation} right). This simulates the real-world situation where the environment does not delay or accelerate according to the agent's computation time.

\noindent\textbf{Games}
In order to control the dynamics of the environment for evaluating real-time reasoning, we use real-time games
in our gym. We created three games to capture different challenges that a dynamic environment brings: maintaining safety when hazards happen, seizing transient opportunities, and coordinating with independent partners (Tab. \ref{tab:cognitive_load}).
\begin{table}[!b]
    \centering
    \small
    \caption{Different Games in \ourBenchmark.}
    \begin{tabular}{llll}
        \toprule \textbf{Game} & \textbf{Dynamic Aspect} & \textbf{Cognitive Load Factor} & \textbf{Evaluation Metrics}       \\
        \midrule Freeway   & Hazards            & Min steps to finish: $S$ & \#Steps the agent takes to get to the other side \\
        Snake           & Opportunities              & \#Obstacles: $N$           & \#Apples the agent eats before collision   \\
        Overcooked     & Partners               & Kitchen Counter Len.: $L$   & \#Orders completed cooperatively             \\
        \bottomrule
    \end{tabular}
    \label{tab:cognitive_load}
\end{table}

In Freeway, the agent traverses multiple lanes of bidirectional traffic, which requires constant monitoring for oncoming cars while planning future trajectories to avoid becoming trapped mid-road.
%agents to actively monitor new cars coming in both sides while planning future trajectories to avoid becoming trapped mid-road. 
In Snake, the agent eats apples which are only available for a short period of time, creating opportunities that must be seized quickly, while greedy food collection might lead to positions where the growing snake traps itself. 
In Overcooked, the agent collaborates with a scripted partner following a non-stationary policy (App. \S~\ref{sec:env_details}). Efficient dish preparation requires not only planning a sequence of actions, but also coordinating effectively with the partner.

\noindent\textbf{Cognitive Load} 
To systematically control how challenging the games are, we make the difficulty of each game tunable through a cognitive load factor (Tab. \ref{tab:cognitive_load}). In Freeway, difficulty is determined by the minimum number of steps required to traverse the road,
since longer paths typically introduce more detours and require deeper planning horizon. 
In Snake, we vary the density of obstacles, increasing route complexity and the need for look-ahead. 
In Overcooked, complexity is controlled by the length of an internal kitchen counter, as a longer counter expands navigation complexity and stretches temporal windows for high-level goals, creating larger discrepancies in long-term planning and immediate execution. 
For each game, we design 3 levels of difficulties, easy, medium, and hard; the corresponding ranges of each level can be found in App. Tab. \ref{tab:game_parameters}. 

\begin{figure}[!t]
    \centering
    \includegraphics[width=1\linewidth]{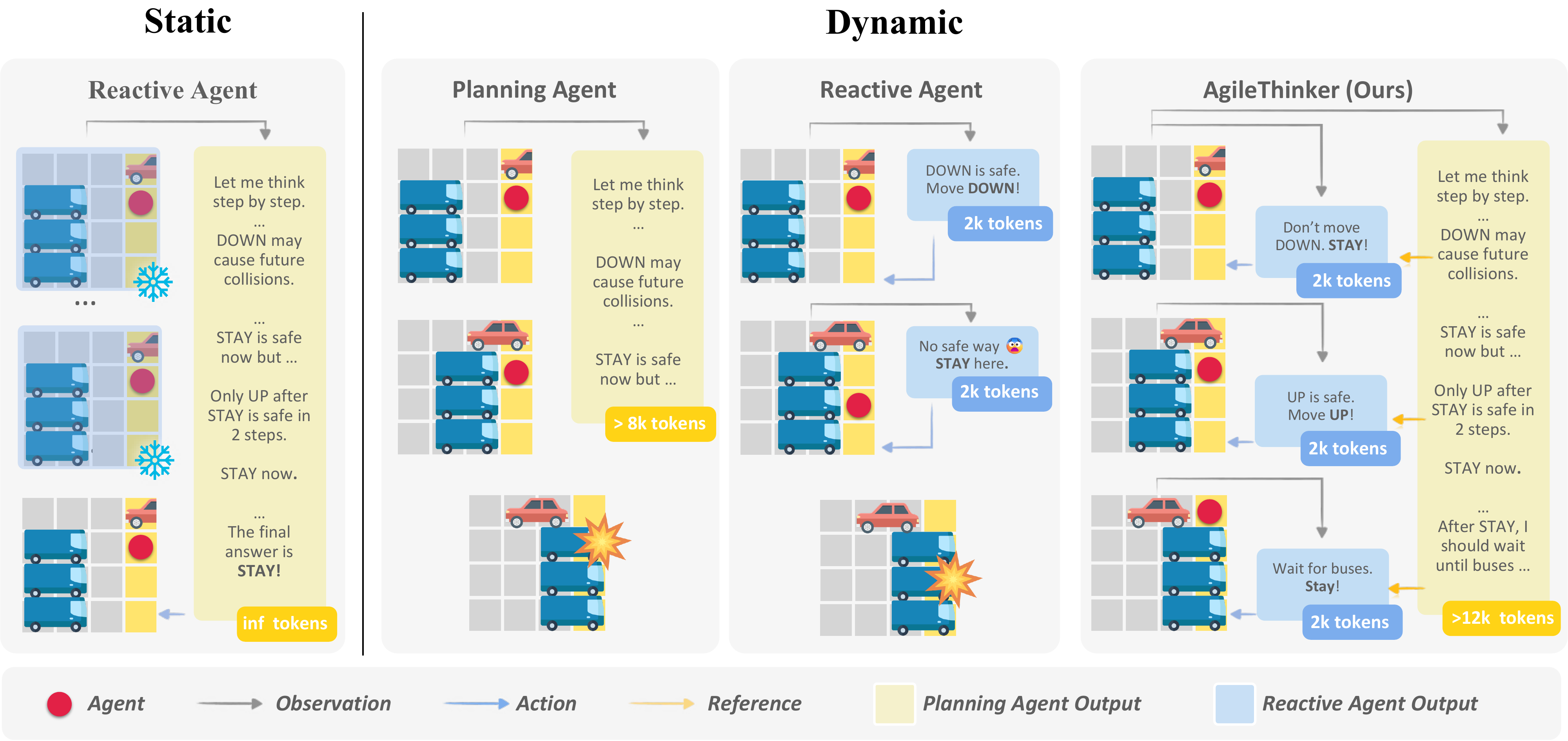}
    \caption{Existing evaluation setups for LLM Agents often assume a \emph{static}
    setting, where the environment halts while the agent completes reasoning
    with unlimited computation. In \ourBenchmark, environments are \emph{dynamic},
    evolving regardless of agents' computation state. As illustrated in the
    \emph{Freeway} setting, \textbf{\textcolor{slow}{Planning Agent}}, which
    performs extended reasoning without interruption, fails to act timely while \textbf{\textcolor{fast}{Reactive
    Agent}}, which performs reasoning strictly within environment update period,
    lacks foresight and collides. \textbf{\textcolor{parallel}{{\ourSystem}}}\space
    combines both timely reaction and long-term planning to navigate such environments
    effectively.}
    \label{fig:envs}
\end{figure}

\paragraph{Time Pressure} 
To enable reproducible and hardware-agnostic evaluation, we use token count as a time proxy to simulate the games, leveraging the fact that LLM decoding time scales almost linearly with output length through time-per-output-token (TPOT), while prefilling time becomes negligible for long sequences. This yields decoding time $T = N_{T} \times \text{TPOT}$, where $N_{T}$ is the generated token count, allowing fair comparison across deployment scenarios. We impose time pressure by letting the environment step every $N_{T_{\mathcal{E}}}$, or \texttt{T\_E} in Fig. \ref{fig:formulation}, tokens generated by agents.
When the agent cannot produce a valid action, we let the environment step with a \texttt{DEFAULT\_ACTION} (Fig. \ref{fig:formulation}). In Freeway and Snake, the default action is moving in the same direction as before, and in Overcooked, the default action is to stay idle. We consider four different time pressure levels, 32k, 16k, 8k, and 4k tokens per step, from low to high pressure. As shown in Figure \ref{fig:envs}, unlike existing static evaluations, the introduction of time pressure simulates the real world dynamic environments where the world does not freeze during agent reasoning.

\paragraph{Evaluation} Evaluation metric differs for each game. In Freeway, we evaluate the number of steps the agent takes to get to the other side, while the agent is reset to the origin every time it gets hit by a car; in Snake, we count the number of apples that the agent eats before a collision; and in Overcooked, we use the number of orders that the agent and the partner completed in total. These evaluation metrics represent the capability of the agents to solve the tasks not only logically but also timely. For each game, we normalize the scores by the highest score the agent could get in that game, so we always have a score between 0 and 1. As cognitive load and time pressure increase, we expect the scores decrease. However, the scores of an agent with strong real-time reasoning capabilities should decrease slower. It is worth noting that our gym is used to evaluate \emph{design choices of agent systems} when the model or model family (e.g. DeepSeek-V3 and R1) is fixed. Therefore, cross-model comparisons may be unfair due to their different tokenizers and underlying architectures.

    \section{Real-time Reasoning Agents}
\label{sec:real-time-reasoning-agents}

To address the real-time reasoning problem, we consider two solution paradigms: (1) reactive, where the agent produces a new action at every environment step, and (2) planning, where the agent reasons across multiple steps to generate an action plan, which is then executed until the agent resumes reasoning. In the following, we discuss how to create agents following each of the two paradigms, and how these two paradigms are engaged in \ourSystem.% As illustrated in Fig. \ref{fig:envs}, planning agents produce slower and more deliberate reasoning, reactive agents provide faster and more intuitive reasoning, while \ourSystem leverage both systems to handle real-time reasoning. 

\paragraph{Reactive agents} We constrain reactive agents by a token budget $N_i$, ensuring they can respond within each environment update when $N_i \leq N_{T_{\mathcal{E}}}$. We consider two kinds of language models for reactive agents: (1) non-thinking models\footnote{Thinking models are the LLMs trained with reinforcement learning to incentivize reasoning before generating answers \citep{deepseekai2025deepseekr1incentivizingreasoningcapability}, while non-thinking models are the LLMs that have not been specifically trained to generate long reasoning.} that produce limited tokens for each response; and (2) thinking models that produce extended reasoning which is cut off at the token budget $N_i$ with budget forcing \citep{muennighoff2025s1simpletesttimescaling}. In both cases, the agent produces one action per environment step, enabling immediate reaction to any change. This reactive approach is commonly used in agent systems these days. 

\paragraph{Planning agents} %Reactive agents do not have the resources to perform reasoning that would take more than one environment step, which restricts their capability of dealing with complex situations.
While reactive agents ensure timeliness, their token budget restricts the complexity of reasoning they can perform within a single step. 
In contrast, planning agents can consider their plans through multiple environment steps. We consider two formats of plans: (1) multi-step actions where a thinking model is used to generate a sequence of multiple actions to be executed; and (2) code plans where a thinking model is used to generate a code snippet that automatically produces actions based on observation input \citep{codeaspolicies2022,zhang2025leveragingdualprocesstheory}. Although (1) is often easier to generate, (2) is more adaptive to potential changes. Both formats allow for more deliberate, long-horizon decision-making by leveraging extended reasoning.

\paragraph{\ourSystem}

% \begin{wrapfigure}[19]{r}{0.65\linewidth}
%     \centering
%     \vspace{-13pt}
%     \includegraphics[width=\linewidth]{figures/flow.pdf}
%     \vspace{-8pt}
%     \caption{ Two parallel threads in \ourSystem}
%     \label{fig:pipeline}
% \end{wrapfigure}
\begin{figure*}[t]
    \centering
    \includegraphics[width=\linewidth]{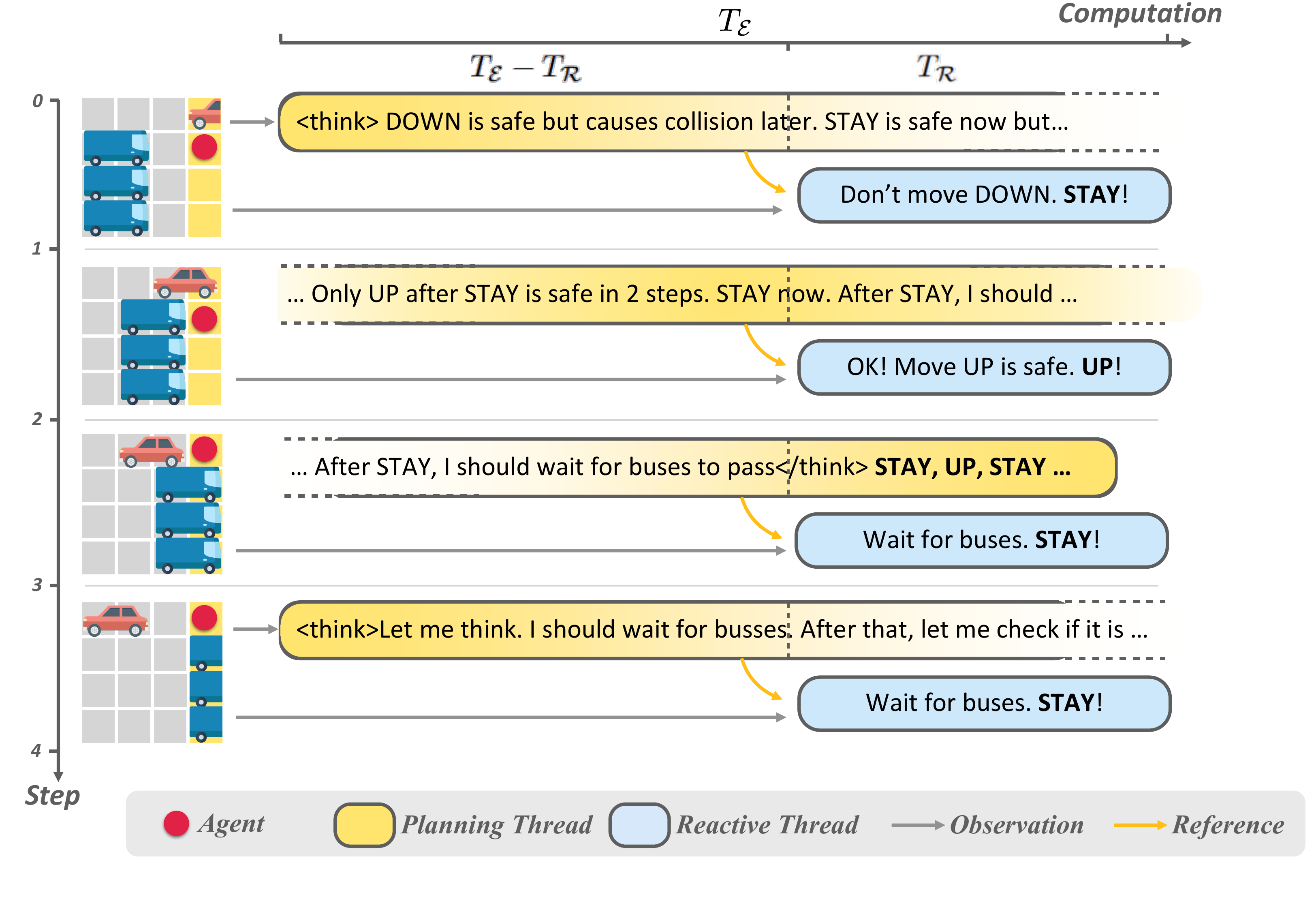}
    \caption{Two parallel threads in \ourSystem}
    \label{fig:pipeline}
\end{figure*}

All agents introduced above must complete their reasoning process before taking any action. To overcome this limitation, we propose \ourSystem, which employs two parallel threads to achieve both timely action generation and uninterrupted deep planning. The planning thread $\mathcal{P}$ runs an LLM that streams the thinking process for a multi-step action plan. Upon initialization, a reasoning process begins that continues until the execution of a plan. $\mathcal{P}$ cannot keep up with environmental changes (e.g. Steps 1-2 in Fig.~\ref{fig:pipeline}). However, since its planning focuses on long-term objectives, many generated insights remain useful over extended horizons. In contrast, the reactive thread $\mathcal{R}$ runs a separate LLM under strict time constraints $T_{\mathcal{R}}\leq T_{\mathcal{E}}$, making decisions based on the latest observation and the (partial) output of $\mathcal{P}$ (see gray and yellow arrows in Fig.~\ref{fig:pipeline}). 

The coordination between the two threads $\mathcal{R}$ and $\mathcal{P}$ follows a time-sharing protocol: during each environment step, $\mathcal{P}$ operates continuously while $\mathcal{R}$ activates only in the final $T_{\mathcal{R}}$ time units. The hyperparameter $T_{\mathcal{R}}$ controls the resource trade-off between the two thread. With a larger $T_{\mathcal{R}}$, the reactive thread can be more adaptive, but there will be less reasoning from the planning thread to refer to. Effectively balancing planning and reaction resources is the key to success in \ourSystem, which will be discussed in \S\ref{sec:time_management}.

\begin{figure}[!t]
    \centering
    \includegraphics[width=0.95\linewidth]{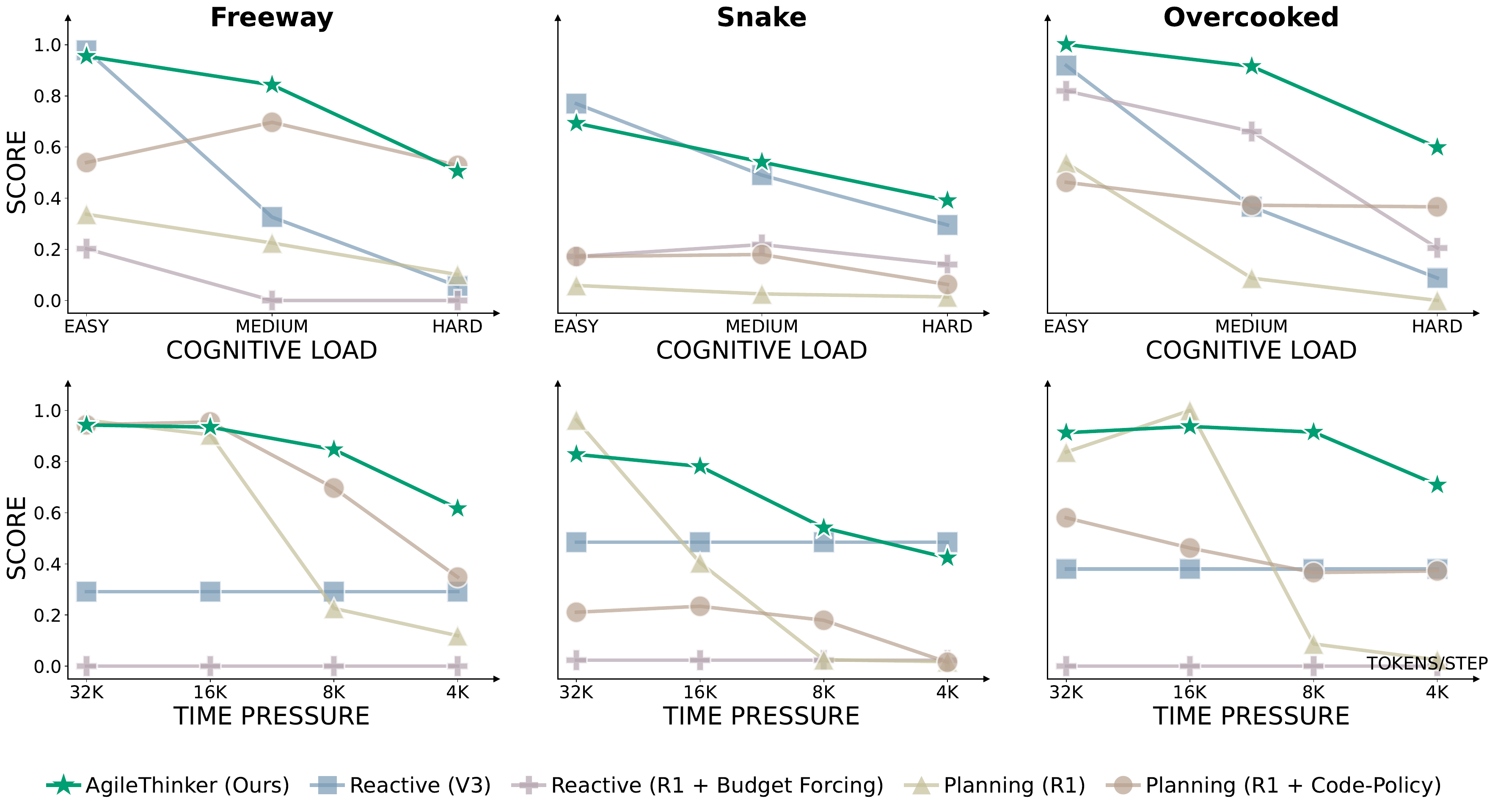}
    \caption{Performance of reasoning agents in \ourBenchmark\space under varying cognitive
    loads and time pressures. Upper: we fix time pressure at 8k tokens per step
    and vary cognitive load. Lower: we fix cognitive load at medium level
    and vary time pressure. Full data and significance test at App. \S~\ref{sec:main_results} and \S~\ref{sec:significance_test}.}
    \label{fig:main_results}
    \vspace{-20pt}
\end{figure}
\section{Is Single Paradigm Enough for Real-Time Reasoning?}

\textbf{Evaluation Setup:} To investigate how \textbf{cognitive load}
and \textbf{time pressure} affect the performance respectively, we conduct two
series of experiments.
(1) Cognitive load varies (Easy, Medium, Hard) while time pressure is fixed at $8\text{k}$ tokens/step---lenient enough for non-thinking models to complete their responses, yet restrictive for thinking models. The intrinsic bound $N_i$ (see Section~\ref{sec:real-time-reasoning-agents}) for reactive agent is set to $8$k.
(2) Time pressure varies ($N_{T_\mathcal{E}}\in \{4\text{k}, 8\text{k}, 16\text{k}, 32\text{k}\}$) with medium cognitive load. Here, $N_i$ is set to $4$k to ensure it remains lower than time pressure budget. We evaluate each agent 32 times (8 game seeds $\times$ 4 LLM sampling seeds) under each setting and report the average score of these samples. Details of the environments and score calculation can be found in
App. \S~\ref{sec:env_details}, and prompts are provided in App. \S~\ref{sec:prompts}.

\begin{figure}[!t]
    \centering
    \includegraphics[width=0.97\linewidth]{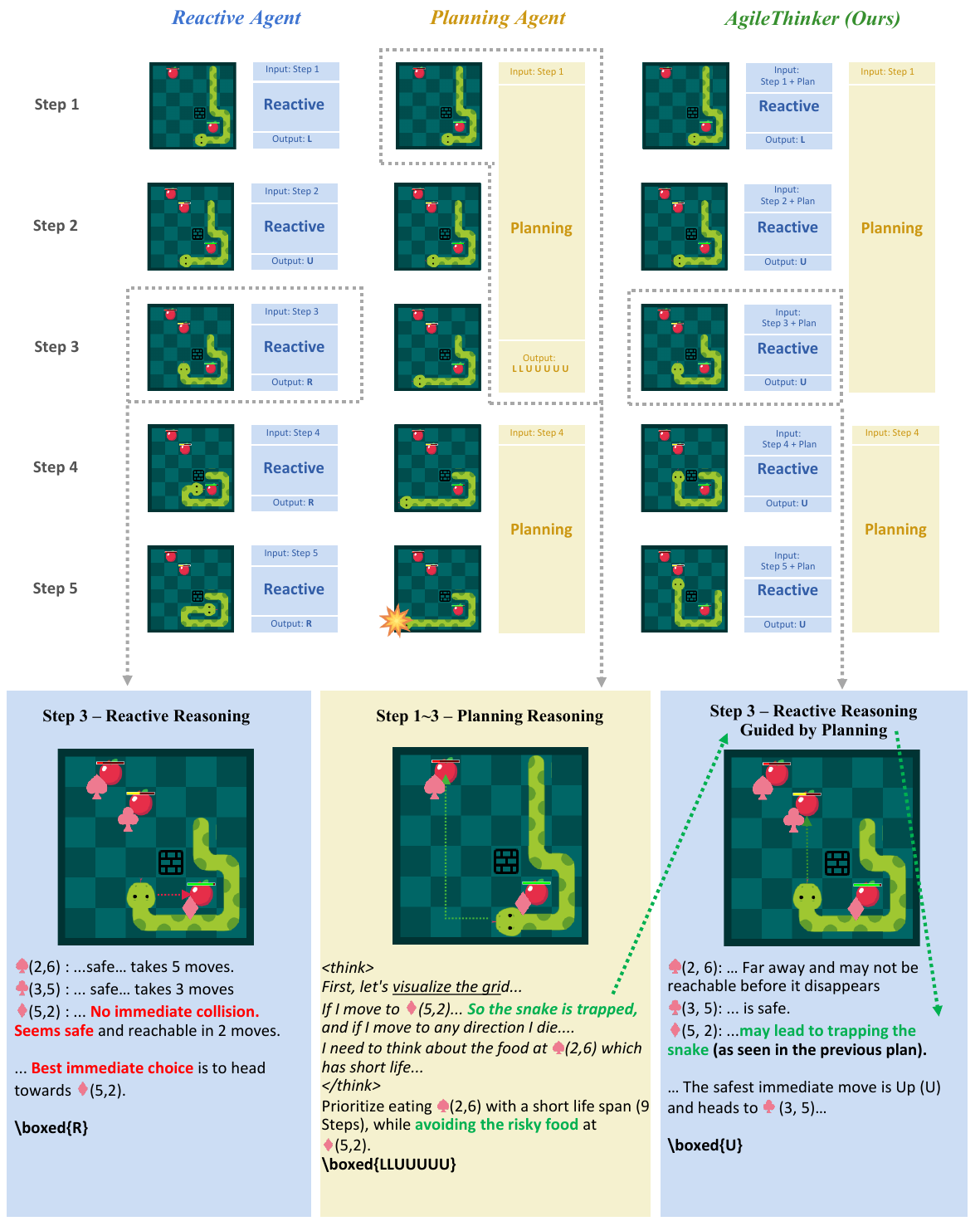}
    \caption{
    \textbf{Thinking trajectories of different paradigms at critical
    steps} At step 3, \textcolor{fast}{\textbf{Reactive Agent (V3)}} greedily pursues the nearest food and collides inevitably after three steps. %without detecting the risk, leading to an inevitable collision after three steps. 
    \textcolor{slow}{\textbf{Planning Agent (R1) }},
    still reasoning over the outdated step-1 state, defaults left.
    However, it correctly identifies that eating the nearest food would result
    in a future collision, and that its lifespan is sufficient to delay consumption.
    Guided by the reasoning of Reactive Thread, Planning Thread in the \textcolor{parallel}{\textbf{\ourSystem}}
    anticipates the trap and chooses to move upward toward a safer food target.}
    
    \vspace{-10pt}
    \label{fig:case-study}
\end{figure}

Figure~\ref{fig:teaser} reports the average scores over three games, while Figure~\ref{fig:main_results} provides a breakdown per game. The results show that reactive and planning agents fail to balance decision quality and efficiency, whereas our \ourSystem achieves robust performance under varying conditions.

\textbf{Reactive agent sacrifices decision quality for efficiency.} By design, the reactive agent restricts computation time less than $T_{\mathcal{E}}$ and maintains consistent performance across all time pressures. However, the limit on test-time scaling also causes a dramatic performance drop as cognitive load increases (scores falling from 0.89 to 0.15, versus 0.88 to 0.50 for \ourSystem)
This drop stems from its inability to consider future consequences of a move carefully. As exemplified in the case study in Figure~\ref{fig:case-study}, the reactive agent greedily pursues immediate rewards, falling into predictable traps while \ourSystem\space avoids by considering long-term survival requirements.

\textbf{Planning agent optimizes for decision quality but suffers under time pressure.}
Planning agent excels under relaxed time constraints but suffers from dramatic degradation when time pressure increases (scores dropping from 0.92 to 0.05, versus 0.90 to 0.58 for \ourSystem). 
Its fundamental flaw is obliviousness to environmental changes, executing plans based on outdated observations.
As illustrated in the case study, the agent is unaware that the snake has moved forward during reasoning, hence it crashes into a wall by following the obsolete plan. However, reactive thread in \ourSystem\space is informed of the latest state, thus able to adjust the output of planning thread.

The two variants of reactive and planning agents discussed in \S\ref{sec:real-time-reasoning-agents} are also insufficient. Neither method matches reactive agent's performance under time constraints while maintaining planning agent's performance under relaxed conditions.
R1 with budget forcing fails to reliably shorten generation length without harming quality, yielding even worse performance than the reactive agent based on V3 ($0.01 < 0.39$). Experiment trajectories reveal that meaningful actions mainly occur when generation naturally fits within the budget; forced truncation typically results in no-ops. More advanced budget control methods, such as budget-aware generation (Figure~\ref{fig:token_usage_vs_budget}), only offer minor shifts in token usage distribution, and truncation still leads to severe performance drops.
R1 with code-policy helps in tasks with simple algorithmic solution like Freeway. However, it underperforms direct action generation in the other games, where effective policies require complex contextual understanding capabilities like heuristic search or Theory-of-Mind (inferring partner's mental state through action history) that cannot be easily compressed into code.
To better illustrate this, we show representative code policies generated by R1 in App. \S\ref{sec:code_as_policy}.

\section{How to Manage Resources Between Reaction and Planning?}
\label{sec:time_management}

\begin{figure}
    \centering
    \includegraphics[width=0.8\linewidth]{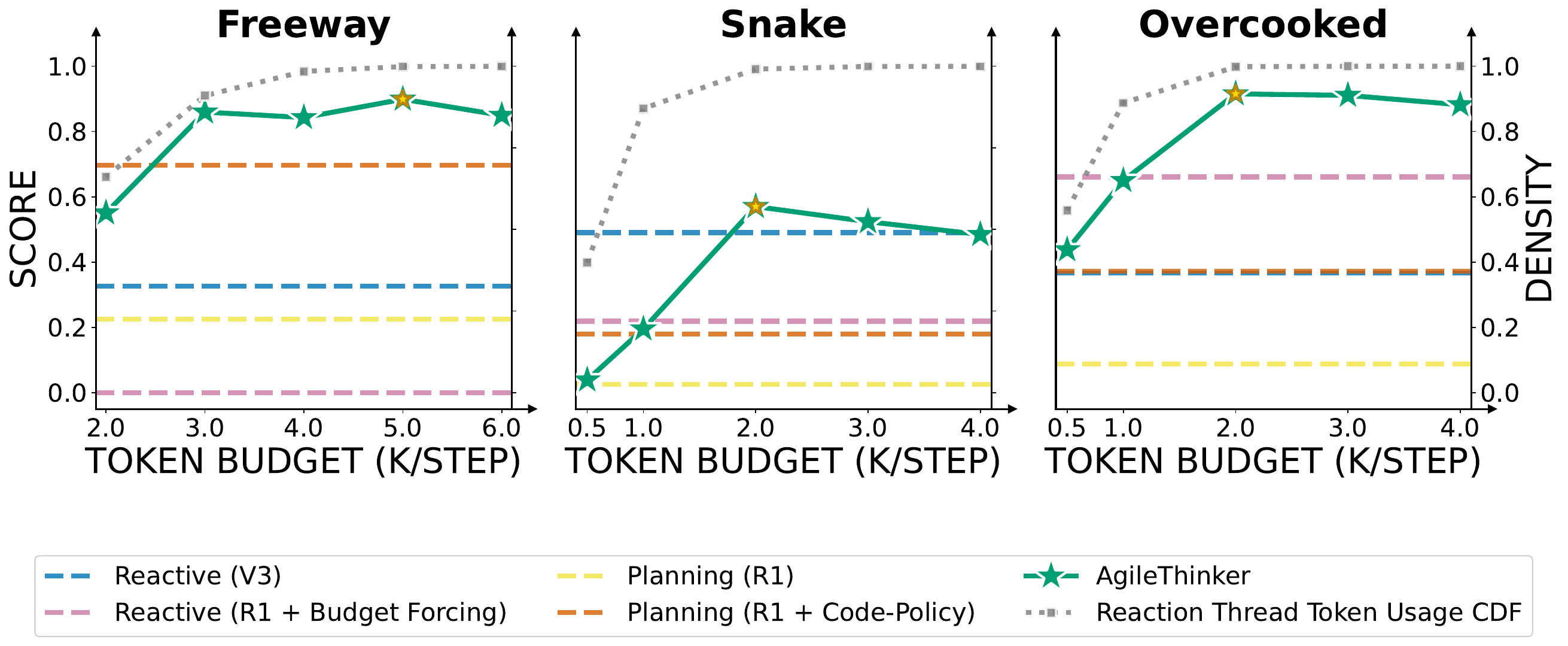}
    \caption{Performance of \ourSystem\space under different reactive thread token budgets $N_{T_{\mathcal{R}}}$. The cumulative distribution function (CDF) shows the natural token usage of $\mathcal{R}$ across all game trajectories when generation is not truncated, indicating inherent computational requirements of $\mathcal{R}$.}
    \label{fig:time_management}
\end{figure}

Effective coordination between reactive thread $\mathcal{R}$ and planning thread $\mathcal{P}$ requires careful time management to determine when to invoke $\mathcal{R}$ within each environment step. We analyze how the token budget $N_{T_{\mathcal{R}}}$ allocated to $\mathcal{R}$ affects the performance of \ourSystem. We set $N_{T_{\mathcal{E}}}$ to 8k and vary the token budget $N_{T_\mathcal{R}}$ from 0.5k to 8k tokens. Figure~\ref{fig:time_management} presents agent scores under different $N_{T_\mathcal{R}}$, where we also plot the cumulative distribution function (CDF) of $\mathcal{R}$'s token usage across all game trajectories without constraints to understand its inherent computational requirements.

%\wenyl{Number, gird \& filling }

We can see that setting $N_{T_\mathcal{R}}$ too small (e.g., 0.5k) leads to low scores, as $\mathcal{R}$ doesn't have enough time to process strategic
guidance from $\mathcal{P}$ and generate well-reasoned actions. Conversely,
setting $N_{T_\mathcal{R}}$ too large creates idle periods where $\mathcal{R}$
has completed action generation but $\mathcal{P}$ continues productive reasoning.
Empirically, performance peaks when $N_{T_{\mathcal{R}}}$ approximates the natural token upper bound of $\mathcal{R}$, as indicated by the CDF of $\mathcal{R}$'s token usage. This suggests that
$\mathcal{R}$ benefits from fully utilizing its allocated time without truncation or extended idling.

It is worth noting that the optimal time budget varies across environments and
requires empirical tuning. In \textit{Freeway}, optimal budget is approximately 5k
tokens, while \textit{Snake} and \textit{Overcooked} achieve peak performance
around 2k tokens. However, \ourSystem\space consistently outperform single-system
baselines across broad budget ranges, suggesting that rough upperbound estimations
are sufficient to achieve the advantages.

\section{Performance Improvement Under Wall-Clock Time}

\begin{wraptable}{r}{0.65\linewidth}

    \centering
    \small\vspace{-10pt}
    \caption{Wall-clock time performance comparison across agent systems, confirming
    \ourSystem\space advantages persist in real-world deployment scenarios.}
    \begin{tabular}{llll}
        \toprule \textbf{Environment} & \textbf{Reactive (V3) } & \textbf{Planning (R1) } & \textbf{\ourSystem} \\
        \midrule Freeway              & 0.24                  & 0.12                  & \textbf{0.88}                 \\
        Snake                         & 0.37                  & 0.04                  & \textbf{0.45}                 \\
        Overcooked                    & 0.57                  & 0.00                  & \textbf{0.89}                 \\
        \bottomrule
    \end{tabular}
    
    \label{tab:wall_time_score}
\end{wraptable}

To validate the practical applicability of our token-based simulation, we conduct experiments using actual wall-clock time with official API of DeepSeek. Our results show token count has strong linear correlation with physical inference time. Specifically, we model this relationship as $T = \alpha N + \beta$, where $T$ represents total runtime and $N$ represents generated tokens. Least squares estimation on all experiment trajectories (plotted in Figure~~\ref{fig:wall_time_align}) yields $\alpha = 0.0473$ s/token, $\beta = 334.55$ s, with $R^{2}= 0.9986$. This near-perfect correlation validates our token-based temporal abstraction and
confirms its practical relevance for real-world deployments.

We also conduct experiments to verify that the advantage of \ourSystem\space remains when the
game is simulated in wall-clock time. Using the derived TPOT of $0.047$ s/token,
we evaluate agent systems with environment evolution intervals of $T_{\mathcal{E}}
= 6$ minutes, corresponding to approximately 8,000 tokens per step. Table~\ref{tab:wall_time_score}
shows that \ourSystem\space consistently outperforms both Reactive and Planning
Agents in physical time. These results establish that our framework's benefits extend
beyond theory to practical applications, demonstrating applicability of our
architecture for agent deployments on intelligence-demanding, real-time tasks.

    \section{Related Work}

\paragraph{Evaluation Environments for LLM Agents:} Existing evaluation setups for
LLM agents mostly focus on \emph{static} environments where nothing changes during
episodes \citep{yang2024sweagentagentcomputerinterfacesenable,zhou2024webarenarealisticwebenvironment}
or state pauses during LLM reasoning \citep{zhou2024hazardchallengeembodieddecision, shi2025korgymdynamicgameplatform}.
This unrealistic assumption risks performance drop and even safety hazards when
applying LLM agents in latency-sensitive applications \citep{sinha2024realtime, Zheng2025RapidReflex}.
Prior work has modeled computation delays through Delay-Aware MDPs \citep{chen2020delayawaremodelbasedreinforcementlearning},
sticky-action schemes \citep{mahmood2018benchmarkingreinforcementlearningalgorithms},
and asynchronous interactive MDPs \citep{Travnik_2018, riemer2024enablingrealtimereinforcementlearning},
but the scope of these works is limited to traditional reinforcement learning. Although some works \citep{liu2024llmpoweredhierarchicallanguageagent,zhang2025leveragingdualprocesstheory}
do adopt LLM agents in wall-clock time, 
our work is the first to formalize real-time reasoning problem for LLM Agents. In particular, we measure elapsed token count as a hardware-agnostic temporal unit, enabling fair and reproducible comparison across agent
systems.

\paragraph{Budget Control for Reasoning Models:} Test-time compute improves LLM performance
but significantly increases inference time, with overthinking behaviors commonly observed in current reasoning models \citep{chen2025think23overthinkingo1like}.
Budget control aims to maximize LLM performance under fixed budgets, and popular methods include early truncation \citep{muennighoff2025s1simpletesttimescaling},
prompting \citep{pu2025thoughtterminatorbenchmarkingcalibratingmitigating} and
training \citep{aggarwal2025l1controllinglongreasoning,kimiteam2025kimik15scalingreinforcement,gemini2025gemini25}.
These methods are effective to a certain extent, but still struggle with precise control over generated token count \citep{alomrani2025reasoningbudgetsurveyadaptive} and performance drops when budgets are far from adequate \citep{han2025tokenbudgetawarellmreasoning}.
This suggests that existing techniques are inadequate for handling both loose and tight budget constraints within a single model. 
The results on \ourBenchmark\space demonstrate that SOTA budget control methods cannot effectively balance reaction and planning, necessitating dual LLM architectures for real-time environments.

\paragraph{Dual Agent Systems:} Dual process theory posits that human cognition operates
through two distinct components: \emph{System 1} (fast and intuitive) and \emph{System
2} (slow and deliberate) \citep{evans2013dual, kahneman2011thinking}. This
inspires dual agent system designs, which combine fast modules (finite-state
machines \citep{zhang2025leveragingdualprocesstheory}, vision transformers
\citep{cui2025openhelixshortsurveyempirical}, small language models \citep{liu2024llmpoweredhierarchicallanguageagent}, etc) for \emph{System 1} with
powerful modules (LLMs with tools \citep{christakopoulou2024agentsthinkingfastslow},
LRMs \citep{zhang2025leveragingdualprocesstheory}, etc) for \emph{System 2}.
\ourSystem differs from existing approaches in that \emph{System 1}(Reactive Thread) can access the partial reasoning trace of \emph{System 2}(Planning Thread), allowing informed decision making with minimal delay.

    \section{Conclusion}

In this work, we identified and formalized \textbf{\ourProblem}, a fundamental
challenge faced by agents in real-world deployment. We introduced \textbf{\ourBenchmark},
the first gym for evaluating LLM agents in continuously evolving environments.
It supports independent control of cognitive load and time pressure, using token
count as a hardware-independent temporal measure. Our evaluation revealed
critical shortcomings of existing reasoning paradigms (reactive and planning reasoning). To address this gap,
we proposed \textbf{\ourSystem}, which engages two reasoning paradigms in parallel. Experiments demonstrate
that our method consistently outperforms all baselines, with advantages growing
as cognitive load increases and time constraints tighten. Future work can extend
our gym to more realistic scenarios, improve coordination mechanisms between two threads, or leverage our gym to train urgency-aware LLM agents.

    \section{Limitations}

Our method investigates the real-time reasoning in LLM agents. Although our formulation is general, we only conducted experiments on DeepSeek models due to two reasons: (1) open source models perform poorly in general, thus the difference between different systems is not significant enough, and (2) other commercial model providers, including OpenAI, Google and Anthropic, do not provide reasoning traces which are crucial in our evaluation.
We also try our best to make sure the readers understand that we do not have any empirical evidence showing that the dual system implemented in \ourSystem is precisely modeling human dual systems. The connection and difference require more rigorous evaluation. 
    % \input{sections/llm_use}
    % \input{sections/ethics_statement}
    %    \subsubsection*{Author Contributions}
    %    If you'd like to, you may include a section for author contributions as is done in many journals. This is optional and at the discretion of the authors.

    %    \subsubsection*{Acknowledgments}
    %    Use unnumbered third level headings for the acknowledgments. All acknowledgments, including those to funding agencies, go at the end of the paper.

    \bibliography{reference}

@misc{zhou2024hazardchallengeembodieddecision,
      title={HAZARD Challenge: Embodied Decision Making in Dynamically Changing Environments}, 
      author={Qinhong Zhou and Sunli Chen and Yisong Wang and Haozhe Xu and Weihua Du and Hongxin Zhang and Yilun Du and Joshua B. Tenenbaum and Chuang Gan},
      year={2024},
      eprint={2401.12975},
      archivePrefix={arXiv},
      primaryClass={cs.CV},
      url={https://arxiv.org/abs/2401.12975}, 
}

@misc{sinha2024realtime,
      title={Real-Time Anomaly Detection and Reactive Planning with Large Language Models}, 
      author={Rohan Sinha and Amine Elhafsi and Christopher Agia and Matthew Foutter and Edward Schmerling and Marco Pavone},
      year={2024},
      eprint={2407.08735},
      archivePrefix={arXiv},
      primaryClass={cs.RO}
}

@misc{yao2022react,
      title={ReAct: Synergizing Reasoning and Acting in Language Models}, 
      author={Shunyu Yao and Jeffrey Zhao and Dian Yu and Nan Du and Izhak Shafran and Karthik R. Narasimhan and Yuan Cao},
      year={2022},
      eprint={2210.03629},
      archivePrefix={arXiv},
      primaryClass={cs.LG}
}

@book{kahneman2011thinking,
  title={Thinking, fast and slow},
  author={Kahneman, Daniel},
  year={2011},
  publisher={Farrar, Straus and Giroux}
}

@article{evans2013dual,
  title={Dual-process theories of reasoning: Contemporary issues and developmental applications},
  author={Evans, Jonathan St BT},
  journal={Developmental Review},
  volume={33},
  number={2},
  pages={145--170},
  year={2013},
  publisher={Elsevier}
}

@article{Zheng2025RapidReflex,
  title={LLM-Enhanced Rapid-Reflex Async-Reflect Embodied Agent for Real-Time Decision-Making in Dynamically Changing Environments},
  author={Zheng, Yangqing and Mao, Shunqi and Zhang, Dingxin and Cai, Weidong},
  journal={arXiv preprint arXiv:2506.07223},
  year={2025}
}

@misc{shi2025korgymdynamicgameplatform,
      title={KORGym: A Dynamic Game Platform for LLM Reasoning Evaluation}, 
      author={Jiajun Shi and Jian Yang and Jiaheng Liu and Xingyuan Bu and Jiangjie Chen and Junting Zhou and Kaijing Ma and Zhoufutu Wen and Bingli Wang and Yancheng He and Liang Song and Hualei Zhu and Shilong Li and Xingjian Wang and Wei Zhang and Ruibin Yuan and Yifan Yao and Wenjun Yang and Yunli Wang and Siyuan Fang and Siyu Yuan and Qianyu He and Xiangru Tang and Yingshui Tan and Wangchunshu Zhou and Zhaoxiang Zhang and Zhoujun Li and Wenhao Huang and Ge Zhang},
      year={2025},
      eprint={2505.14552},
      archivePrefix={arXiv},
      primaryClass={cs.CL},
      url={https://arxiv.org/abs/2505.14552}, 
}

@misc{liu2024llmpoweredhierarchicallanguageagent,
      title={LLM-Powered Hierarchical Language Agent for Real-time Human-AI Coordination}, 
      author={Jijia Liu and Chao Yu and Jiaxuan Gao and Yuqing Xie and Qingmin Liao and Yi Wu and Yu Wang},
      year={2024},
      eprint={2312.15224},
      archivePrefix={arXiv},
      primaryClass={cs.AI},
      url={https://arxiv.org/abs/2312.15224}, 
}

@misc{christakopoulou2024agentsthinkingfastslow,
      title={Agents Thinking Fast and Slow: A Talker-Reasoner Architecture}, 
      author={Konstantina Christakopoulou and Shibl Mourad and Maja Matarić},
      year={2024},
      eprint={2410.08328},
      archivePrefix={arXiv},
      primaryClass={cs.AI},
      url={https://arxiv.org/abs/2410.08328}, 
}

@misc{zhang2025leveragingdualprocesstheory,
      title={Leveraging Dual Process Theory in Language Agent Framework for Real-time Simultaneous Human-AI Collaboration}, 
      author={Shao Zhang and Xihuai Wang and Wenhao Zhang and Chaoran Li and Junru Song and Tingyu Li and Lin Qiu and Xuezhi Cao and Xunliang Cai and Wen Yao and Weinan Zhang and Xinbing Wang and Ying Wen},
      year={2025},
      eprint={2502.11882},
      archivePrefix={arXiv},
      primaryClass={cs.AI},
      url={https://arxiv.org/abs/2502.11882}, 
}

@misc{cui2025openhelixshortsurveyempirical,
      title={OpenHelix: A Short Survey, Empirical Analysis, and Open-Source Dual-System VLA Model for Robotic Manipulation}, 
      author={Can Cui and Pengxiang Ding and Wenxuan Song and Shuanghao Bai and Xinyang Tong and Zirui Ge and Runze Suo and Wanqi Zhou and Yang Liu and Bofang Jia and Han Zhao and Siteng Huang and Donglin Wang},
      year={2025},
      eprint={2505.03912},
      archivePrefix={arXiv},
      primaryClass={cs.RO},
      url={https://arxiv.org/abs/2505.03912}, 
}

@misc{chen2020delayawaremodelbasedreinforcementlearning,
      title={Delay-Aware Model-Based Reinforcement Learning for Continuous Control}, 
      author={Baiming Chen and Mengdi Xu and Liang Li and Ding Zhao},
      year={2020},
      eprint={2005.05440},
      archivePrefix={arXiv},
      primaryClass={cs.LG},
      url={https://arxiv.org/abs/2005.05440}, 
}

@misc{mahmood2018benchmarkingreinforcementlearningalgorithms,
      title={Benchmarking Reinforcement Learning Algorithms on Real-World Robots}, 
      author={A. Rupam Mahmood and Dmytro Korenkevych and Gautham Vasan and William Ma and James Bergstra},
      year={2018},
      eprint={1809.07731},
      archivePrefix={arXiv},
      primaryClass={cs.LG},
      url={https://arxiv.org/abs/1809.07731}, 
}

@article{Travnik_2018,
   title={Reactive Reinforcement Learning in Asynchronous Environments},
   volume={5},
   ISSN={2296-9144},
   url={http://dx.doi.org/10.3389/frobt.2018.00079},
   DOI={10.3389/frobt.2018.00079},
   journal={Frontiers in Robotics and AI},
   publisher={Frontiers Media SA},
   author={Travnik, Jaden B. and Mathewson, Kory W. and Sutton, Richard S. and Pilarski, Patrick M.},
   year={2018},
   month=jun }

@misc{riemer2024enablingrealtimereinforcementlearning,
      title={Enabling Realtime Reinforcement Learning at Scale with Staggered Asynchronous Inference}, 
      author={Matthew Riemer and Gopeshh Subbaraj and Glen Berseth and Irina Rish},
      year={2024},
      eprint={2412.14355},
      archivePrefix={arXiv},
      primaryClass={cs.LG},
      url={https://arxiv.org/abs/2412.14355}, 
}

@misc{yang2024sweagentagentcomputerinterfacesenable,
      title={SWE-agent: Agent-Computer Interfaces Enable Automated Software Engineering}, 
      author={John Yang and Carlos E. Jimenez and Alexander Wettig and Kilian Lieret and Shunyu Yao and Karthik Narasimhan and Ofir Press},
      year={2024},
      eprint={2405.15793},
      archivePrefix={arXiv},
      primaryClass={cs.SE},
      url={https://arxiv.org/abs/2405.15793}, 
}

@misc{muennighoff2025s1simpletesttimescaling,
      title={s1: Simple test-time scaling}, 
      author={Niklas Muennighoff and Zitong Yang and Weijia Shi and Xiang Lisa Li and Li Fei-Fei and Hannaneh Hajishirzi and Luke Zettlemoyer and Percy Liang and Emmanuel Candès and Tatsunori Hashimoto},
      year={2025},
      eprint={2501.19393},
      archivePrefix={arXiv},
      primaryClass={cs.CL},
      url={https://arxiv.org/abs/2501.19393}, 
}

@misc{pu2025thoughtterminatorbenchmarkingcalibratingmitigating,
      title={THOUGHTTERMINATOR: Benchmarking, Calibrating, and Mitigating Overthinking in Reasoning Models}, 
      author={Xiao Pu and Michael Saxon and Wenyue Hua and William Yang Wang},
      year={2025},
      eprint={2504.13367},
      archivePrefix={arXiv},
      primaryClass={cs.CL},
      url={https://arxiv.org/abs/2504.13367}, 
}

@misc{gemini2025gemini25,
  title={Gemini 2.5: A Multimodal AI Model},
  author={{Gemini Team}},
  year={2025},
  month={January},
  howpublished={\url{https://storage.googleapis.com/deepmind-media/gemini/gemini_v2_5_report.pdf}},
  note={Technical Report, Google DeepMind}
}

@misc{deepseekai2025deepseekr1incentivizingreasoningcapability,
      title={DeepSeek-R1: Incentivizing Reasoning Capability in LLMs via Reinforcement Learning}, 
      author={DeepSeek-AI and Daya Guo and Dejian Yang and Haowei Zhang and Junxiao Song and Ruoyu Zhang and Runxin Xu and Qihao Zhu and Shirong Ma and Peiyi Wang and Xiao Bi and Xiaokang Zhang and Xingkai Yu and Yu Wu and Z. F. Wu and Zhibin Gou and Zhihong Shao and Zhuoshu Li and Ziyi Gao and Aixin Liu and Bing Xue and Bingxuan Wang and Bochao Wu and Bei Feng and Chengda Lu and Chenggang Zhao and Chengqi Deng and Chenyu Zhang and Chong Ruan and Damai Dai and Deli Chen and Dongjie Ji and Erhang Li and Fangyun Lin and Fucong Dai and Fuli Luo and Guangbo Hao and Guanting Chen and Guowei Li and H. Zhang and Han Bao and Hanwei Xu and Haocheng Wang and Honghui Ding and Huajian Xin and Huazuo Gao and Hui Qu and Hui Li and Jianzhong Guo and Jiashi Li and Jiawei Wang and Jingchang Chen and Jingyang Yuan and Junjie Qiu and Junlong Li and J. L. Cai and Jiaqi Ni and Jian Liang and Jin Chen and Kai Dong and Kai Hu and Kaige Gao and Kang Guan and Kexin Huang and Kuai Yu and Lean Wang and Lecong Zhang and Liang Zhao and Litong Wang and Liyue Zhang and Lei Xu and Leyi Xia and Mingchuan Zhang and Minghua Zhang and Minghui Tang and Meng Li and Miaojun Wang and Mingming Li and Ning Tian and Panpan Huang and Peng Zhang and Qiancheng Wang and Qinyu Chen and Qiushi Du and Ruiqi Ge and Ruisong Zhang and Ruizhe Pan and Runji Wang and R. J. Chen and R. L. Jin and Ruyi Chen and Shanghao Lu and Shangyan Zhou and Shanhuang Chen and Shengfeng Ye and Shiyu Wang and Shuiping Yu and Shunfeng Zhou and Shuting Pan and S. S. Li and Shuang Zhou and Shaoqing Wu and Shengfeng Ye and Tao Yun and Tian Pei and Tianyu Sun and T. Wang and Wangding Zeng and Wanjia Zhao and Wen Liu and Wenfeng Liang and Wenjun Gao and Wenqin Yu and Wentao Zhang and W. L. Xiao and Wei An and Xiaodong Liu and Xiaohan Wang and Xiaokang Chen and Xiaotao Nie and Xin Cheng and Xin Liu and Xin Xie and Xingchao Liu and Xinyu Yang and Xinyuan Li and Xuecheng Su and Xuheng Lin and X. Q. Li and Xiangyue Jin and Xiaojin Shen and Xiaosha Chen and Xiaowen Sun and Xiaoxiang Wang and Xinnan Song and Xinyi Zhou and Xianzu Wang and Xinxia Shan and Y. K. Li and Y. Q. Wang and Y. X. Wei and Yang Zhang and Yanhong Xu and Yao Li and Yao Zhao and Yaofeng Sun and Yaohui Wang and Yi Yu and Yichao Zhang and Yifan Shi and Yiliang Xiong and Ying He and Yishi Piao and Yisong Wang and Yixuan Tan and Yiyang Ma and Yiyuan Liu and Yongqiang Guo and Yuan Ou and Yuduan Wang and Yue Gong and Yuheng Zou and Yujia He and Yunfan Xiong and Yuxiang Luo and Yuxiang You and Yuxuan Liu and Yuyang Zhou and Y. X. Zhu and Yanhong Xu and Yanping Huang and Yaohui Li and Yi Zheng and Yuchen Zhu and Yunxian Ma and Ying Tang and Yukun Zha and Yuting Yan and Z. Z. Ren and Zehui Ren and Zhangli Sha and Zhe Fu and Zhean Xu and Zhenda Xie and Zhengyan Zhang and Zhewen Hao and Zhicheng Ma and Zhigang Yan and Zhiyu Wu and Zihui Gu and Zijia Zhu and Zijun Liu and Zilin Li and Ziwei Xie and Ziyang Song and Zizheng Pan and Zhen Huang and Zhipeng Xu and Zhongyu Zhang and Zhen Zhang},
      year={2025},
      eprint={2501.12948},
      archivePrefix={arXiv},
      primaryClass={cs.CL},
      url={https://arxiv.org/abs/2501.12948}, 
}

@article{Stanovich_West_2000, title={Individual differences in reasoning: Implications for the rationality debate?}, volume={23}, DOI={10.1017/S0140525X00003435}, number={5}, journal={Behavioral and Brain Sciences}, author={Stanovich, Keith E. and West, Richard F.}, year={2000}, pages={645–665}}

@article{doi:10.1177/1745691612460685,
author = {Jonathan St. B. T. Evans and Keith E. Stanovich},
title ={Dual-Process Theories of Higher Cognition: Advancing the Debate},

journal = {Perspectives on Psychological Science},
volume = {8},
number = {3},
pages = {223-241},
year = {2013},
doi = {10.1177/1745691612460685},
    note ={PMID: 26172965},

URL = { 
    
        https://doi.org/10.1177/1745691612460685
    
    

},
eprint = { 
    
        https://doi.org/10.1177/1745691612460685
    
    

}
}

@misc{alomrani2025reasoningbudgetsurveyadaptive,
      title={Reasoning on a Budget: A Survey of Adaptive and Controllable Test-Time Compute in LLMs}, 
      author={Mohammad Ali Alomrani and Yingxue Zhang and Derek Li and Qianyi Sun and Soumyasundar Pal and Zhanguang Zhang and Yaochen Hu and Rohan Deepak Ajwani and Antonios Valkanas and Raika Karimi and Peng Cheng and Yunzhou Wang and Pengyi Liao and Hanrui Huang and Bin Wang and Jianye Hao and Mark Coates},
      year={2025},
      eprint={2507.02076},
      archivePrefix={arXiv},
      primaryClass={cs.AI},
      url={https://arxiv.org/abs/2507.02076}, 
}

@misc{kimiteam2025kimik15scalingreinforcement,
      title={Kimi k1.5: Scaling Reinforcement Learning with LLMs}, 
      author={Kimi Team and Angang Du and Bofei Gao and Bowei Xing and Changjiu Jiang and Cheng Chen and Cheng Li and Chenjun Xiao and Chenzhuang Du and Chonghua Liao and Chuning Tang and Congcong Wang and Dehao Zhang and Enming Yuan and Enzhe Lu and Fengxiang Tang and Flood Sung and Guangda Wei and Guokun Lai and Haiqing Guo and Han Zhu and Hao Ding and Hao Hu and Hao Yang and Hao Zhang and Haotian Yao and Haotian Zhao and Haoyu Lu and Haoze Li and Haozhen Yu and Hongcheng Gao and Huabin Zheng and Huan Yuan and Jia Chen and Jianhang Guo and Jianlin Su and Jianzhou Wang and Jie Zhao and Jin Zhang and Jingyuan Liu and Junjie Yan and Junyan Wu and Lidong Shi and Ling Ye and Longhui Yu and Mengnan Dong and Neo Zhang and Ningchen Ma and Qiwei Pan and Qucheng Gong and Shaowei Liu and Shengling Ma and Shupeng Wei and Sihan Cao and Siying Huang and Tao Jiang and Weihao Gao and Weimin Xiong and Weiran He and Weixiao Huang and Weixin Xu and Wenhao Wu and Wenyang He and Xianghui Wei and Xianqing Jia and Xingzhe Wu and Xinran Xu and Xinxing Zu and Xinyu Zhou and Xuehai Pan and Y. Charles and Yang Li and Yangyang Hu and Yangyang Liu and Yanru Chen and Yejie Wang and Yibo Liu and Yidao Qin and Yifeng Liu and Ying Yang and Yiping Bao and Yulun Du and Yuxin Wu and Yuzhi Wang and Zaida Zhou and Zhaoji Wang and Zhaowei Li and Zhen Zhu and Zheng Zhang and Zhexu Wang and Zhilin Yang and Zhiqi Huang and Zihao Huang and Ziyao Xu and Zonghan Yang and Zongyu Lin},
      year={2025},
      eprint={2501.12599},
      archivePrefix={arXiv},
      primaryClass={cs.AI},
      url={https://arxiv.org/abs/2501.12599}, 
}

@misc{zhou2024webarenarealisticwebenvironment,
      title={WebArena: A Realistic Web Environment for Building Autonomous Agents}, 
      author={Shuyan Zhou and Frank F. Xu and Hao Zhu and Xuhui Zhou and Robert Lo and Abishek Sridhar and Xianyi Cheng and Tianyue Ou and Yonatan Bisk and Daniel Fried and Uri Alon and Graham Neubig},
      year={2024},
      eprint={2307.13854},
      archivePrefix={arXiv},
      primaryClass={cs.AI},
      url={https://arxiv.org/abs/2307.13854}, 
}

@misc{chen2025think23overthinkingo1like,
      title={Do NOT Think That Much for 2+3=? On the Overthinking of o1-Like LLMs}, 
      author={Xingyu Chen and Jiahao Xu and Tian Liang and Zhiwei He and Jianhui Pang and Dian Yu and Linfeng Song and Qiuzhi Liu and Mengfei Zhou and Zhuosheng Zhang and Rui Wang and Zhaopeng Tu and Haitao Mi and Dong Yu},
      year={2025},
      eprint={2412.21187},
      archivePrefix={arXiv},
      primaryClass={cs.CL},
      url={https://arxiv.org/abs/2412.21187}, 
}

@misc{han2025tokenbudgetawarellmreasoning,
      title={Token-Budget-Aware LLM Reasoning}, 
      author={Tingxu Han and Zhenting Wang and Chunrong Fang and Shiyu Zhao and Shiqing Ma and Zhenyu Chen},
      year={2025},
      eprint={2412.18547},
      archivePrefix={arXiv},
      primaryClass={cs.CL},
      url={https://arxiv.org/abs/2412.18547}, 
}

@misc{aggarwal2025l1controllinglongreasoning,
      title={L1: Controlling How Long A Reasoning Model Thinks With Reinforcement Learning}, 
      author={Pranjal Aggarwal and Sean Welleck},
      year={2025},
      eprint={2503.04697},
      archivePrefix={arXiv},
      primaryClass={cs.CL},
      url={https://arxiv.org/abs/2503.04697}, 
}

@article{putta2024agent,
  title={Agent q: Advanced reasoning and learning for autonomous ai agents},
  author={Putta, Pranav and Mills, Edmund and Garg, Naman and Motwani, Sumeet and Finn, Chelsea and Garg, Divyansh and Rafailov, Rafael},
  journal={arXiv preprint arXiv:2408.07199},
  year={2024}
}

@article{ferrag2025llm,
  title={From llm reasoning to autonomous ai agents: A comprehensive review},
  author={Ferrag, Mohamed Amine and Tihanyi, Norbert and Debbah, Merouane},
  journal={arXiv preprint arXiv:2504.19678},
  year={2025}
}

@article{gou2023tora,
  title={Tora: A tool-integrated reasoning agent for mathematical problem solving},
  author={Gou, Zhibin and Shao, Zhihong and Gong, Yeyun and Shen, Yelong and Yang, Yujiu and Huang, Minlie and Duan, Nan and Chen, Weizhu},
  journal={arXiv preprint arXiv:2309.17452},
  year={2023}
}

@misc{openaigym,
  Author = {Greg Brockman and Vicki Cheung and Ludwig Pettersson and Jonas Schneider and John Schulman and Jie Tang and Wojciech Zaremba},
  Title = {OpenAI Gym},
  Year = {2016},
  Eprint = {arXiv:1606.01540},
}

@inproceedings{codeaspolicies2022,
    title={Code as Policies: Language Model Programs for Embodied Control},
    author={Jacky Liang and Wenlong Huang and Fei Xia and Peng Xu and Karol Hausman and Brian Ichter and Pete Florence and Andy Zeng},
    booktitle={arXiv preprint arXiv:2209.07753},
    year={2022}
}
    \bibliographystyle{iclr2026_conference}

    \appendix

\section{Environment Details}
\label{sec:env_details}

\begin{itemize}
    \item \textbf{Freeway:} The player navigates across parallel highways with moving cars. At each step the player can move to an adjacent lane or
        stay in place. Cars move forward at constant speeds and new cars may
        spawn on either side of the road. If the player is hit by a car, it will be reset to the starting position. The game terminates if the player crosses the road or if the step limit $M = 100$ is reached. The reward for a trajectory $\tau$ is computed as:
        \[
            R(\tau) = M - |\tau|
        \]

    \item \textbf{Snake:} The player controls a snake in a 2D rectangular grid
        with surrounding walls and internal obstacles. At each step, the snake
        can move one step left, right, up or down. If the snake head collides with
        an obstacle or its body segment, it dies. Foods spawn continuously in the
        map and disappears after a fixed number of steps; eating food increases
        the snake's length by 1 unit. The game terminates if the snake dies or the
        number of steps exceeds threshold $M = 100$. The reward is calculated as:
        \[
            R(\tau) = \text{Number of eaten food}- \mathbb{I}[\text{Dies in }M \text{
            steps}]
        \]

    \item \textbf{Overcooked:} A fully observable two-player cooperative game
        where players must collect onions, cook them in a pot and serve the cooked
        soup for rewards. At each step, the players can move in 4 directions or use
        interact action to trigger some events, such as picking or placing an
        item depending on the game state. Since we focus on single-agent settings,
        we model the second player as part of the changing environment, controlled
        by a manually written script for simplicity and consistency. This agent
        randomly chooses one policy to follow: \emph{deliver an onion into an arbitrary pot or a kitchen counter}. The game runs for $M = 100$ steps, and rewards
        are assigned for accomplishment of special events listed in Table \ref{tab:overcooked_rewards}.
        The game is implemented based on the repository \url{https://github.com/HumanCompatibleAI/overcooked_ai}
        \begin{table}[H]
            \begin{center}
                \begin{tabular}{ll}
                    \toprule \bf Event                            & \bf Reward \\
                    \midrule Picking up a dish from the dispenser & 3          \\
                    Picking up a cooked soup from the pot         & 5          \\
                    Serving the soup                              & 20         \\
                    \bottomrule
                \end{tabular}
            \end{center}
            \caption{Rewards for different events in the Overcooked environment.}
            \label{tab:overcooked_rewards}
        \end{table}
\end{itemize}

The reward is then normalized to $[0, 1]$ to get the final game score.
Specifically, let $R_{\text{min}}$ and $R_{\text{max}}$ be the minimum and maximum
rewards observed in all trajectories. The score $S$ is computed as: $S = \frac{R
- R_{\text{min}}}{R_{\text{max}}- R_{\text{min}}}$ Empirically, the
$R_{\text{max}}$ and $R_{\text{min}}$ of each environment are listed in Table
\ref{tab:reward_range}.

\begin{table}[H]
    \begin{center}
        \begin{tabular}{lll}
            \toprule \textbf{Environment} & $R_{\text{min}}$ & $R_{\text{max}}$ \\
            \midrule Freeway              & 0                & 89               \\
            Snake                         & -1               & 15               \\
            Overcooked                    & 0                & 56               \\
            \bottomrule
        \end{tabular}
    \end{center}
    \caption{Minimum and maximum rewards for each environment.}
    \label{tab:reward_range}
\end{table}
\begin{table}[!h]
    \centering
    \small
    \caption{Game difficulty settings, showing ranges for easy, medium, and hard levels.}
    \begin{tabular}{llllll}
        \toprule \textbf{Game} & \textbf{Dynamic Aspect} & \textbf{Cognitive Load Factor} & \textbf{Easy} & \textbf{Medium}     & \textbf{Hard}       \\
        \midrule Freeway   & Hazards            & Min steps to finish: $S$ & $S \leq 12$   & $13 \leq S \leq 16$ & $17 \leq S \leq 21$ \\
        Snake           & Opportunities              & \#Obstacles: $N$           & $N = 1$       & $2 \leq N \leq 5$   & $6 \leq N \leq 8$   \\
        Overcooked     & Partners               & Kitchen Counter Len.: $L$    & $L = 0$       & $L = 3$             & $L = 4$             \\
        \bottomrule
    \end{tabular}
    \label{tab:game_parameters}
\end{table}

\section{Prompt}
\label{sec:prompts}

\textbf{Prompt for the Planning Agent}
\newline

\begin{promptbox}
    Now a player is playing a multi-turn game, and suppose current turn is
    $\{t_{1}\}$. Given the initial position $(0, y_{t_1})$ on a 2D grid (vertical
    axis $y = 0, 1, \dots, 9$), determine the minimal number of turns $H$ and a
    sequence of actions $\{a_{t_1 + t}\}_{t=0}^{H-1}$ to reach $(0, 9)$, avoiding
    collisions with cars on freeways $y = 1, \dots, 8$.

    \rule{\textwidth}{0.5pt}

    \textbf{1. Game Dynamics:}

    \begin{itemize}
        \item \textbf{Player update:} $y_{t+1}= y_{t} + \Delta y_{t}$, where
            \[
                \Delta y_{t}=
                \begin{cases}
                    +1 & \text{if }a_{t}= U \\
                    -1 & \text{if }a_{t}= D \\
                    0  & \text{if }a_{t}= S
                \end{cases}, \quad y_{t+1}\in [0,9]
            \]

        \item \textbf{Car update rules:}\\ For car $k$ on freeway $i$, suppose
            its head is at $h$, tail is at $\tau$ at turn $t_{1}$, and speed is
            $s$. Then at turn $T > t_{1}$, the car span becomes:
            \begin{itemize}
                \item Left-moving:
                    $\text{Span}(t_{1}) = [h, \tau] \rightarrow \text{Span}(T) =
                    [h - s (T-t_{1}), \tau - s (T-t_{1})]$

                \item Right-moving:
                    $\text{Span}(t_{1}) = [\tau, h] \rightarrow \text{Span}(T) =
                    [\tau + s (T-t_{1}), h + s (T-t_{1})]$
            \end{itemize}

        \item Collision occurs at turn $T$ only if $0 \in \text{Span}(T)$ for any
            car on freeway $y_{T}$.

        \item Note that if you decide to move to $y_{T+1}\neq y_{T}$ at turn $T$,
            you will \textbf{NOT} be considered to be on $y_{T+1}$ at turn $T$,
            thus will \textbf{NOT} be collided by cars on $y_{T+1}$ if $0 \in \text{Span}
            (T)$ but $0 \notin \text{Span}(T+1)$.
    \end{itemize}

    \rule{\textwidth}{0.5pt}

    \textbf{2. Task (Turn $t_{1}$):}

    Find a sequence of actions $\{a_{t_1 + t}\}_{t=1}^{H-1}$ which minimizes $H$
    such that $y_{t_1 + H - 1}= 9$.

    \textbf{Answer Format}:

    \begin{center}
        \fbox{\begin{minipage}{0.8\textwidth}
            Turn $t_{1}$: $a_{t_1}$\\ Turn $t_{1} + 1$: $a_{t_1 + 1}$\\ $\vdots$
        \end{minipage}}
    \end{center}

    \rule{\textwidth}{0.5pt}

    \textbf{3. Current State (Turn $t_{1}$):}

    \textbf{Current Turn:} $t_{0} = 10$

    \textbf{Player Position:} $(0, 6)$

    \textbf{Car State:}

    \begin{center}
        \begin{tabular}{|c|p{10cm}|}
            \hline
            \textbf{Freeway $k$} & \textbf{Cars (head $h$, tail $\tau$, direction $d$, speed $s$)} \\
            \hline
            1                    & $(48, 37, \text{right}, 12)$, $(0, -11, \text{right}, 12)$      \\
            \hline
            2                    & $(48, 1, \text{right}, 48)$                                     \\
            \hline
            ...                  & ...                                                             \\
            \hline
            8                    & $(48, 37, \text{right}, 12)$, $(0, -11, \text{right}, 12)$      \\
            \hline
        \end{tabular}
    \end{center}
\end{promptbox}

\newpage
\textbf{Prompt for the Reactive Agent}
\newline
\begin{promptbox}
    \footnotesize \textbf{Prompt:}

    You are a player in a freeway game, starting at $(0, y_{t_0})$ on a 2D grid
    (vertical axis $y = 0, 1, \dots, 9$). Your goal is to reach $(0, 9)$ while avoiding
    collisions with cars on freeways $y = 1, \dots, 8$.

    \rule{\textwidth}{0.5pt}

    \textbf{1. Game Dynamics:}

    \begin{itemize}
        \item \textbf{Player update:}\\ $y_{t+1}= y_{t} + \Delta y_{t}$, where
            \[
                \Delta y_{t}=
                \begin{cases}
                    +1 & \text{if }a_{t}= U \\
                    -1 & \text{if }a_{t}= D \\
                    0  & \text{if }a_{t}= S
                \end{cases}, \quad y_{t+1}\in [0,9]
            \]

        \item \textbf{Car update rules:}\\ For car $k$ on freeway $i$, suppose
            its head is at $h$, tail is at $\tau$ at turn $t_{0}$, and speed is
            $s$. Then at turn $T > t_{0}$, the car span becomes:
            \begin{itemize}
                \item Left-moving:
                    $\text{Span}(t_{0}) = [h, \tau] \rightarrow \text{Span}(T) =
                    [h - s (T-t_{0}), \tau - s (T-t_{0})]$

                \item Right-moving:
                    $\text{Span}(t_{0}) = [\tau, h] \rightarrow \text{Span}(T) =
                    [\tau + s (T-t_{0}), h + s (T-t_{0})]$
            \end{itemize}

        \item Collision occurs at turn $T$ only if $0 \in \text{Span}(T)$ for any
            car on freeway $y_{T}$.

        \item Note that if you decide to move to $y_{T+1}\neq y_{T}$ at turn $T$,
            you will \textbf{NOT} be considered to be on $y_{T+1}$ at turn $T$,
            thus will \textbf{NOT} be collided by cars on $y_{T+1}$ if $0 \in \text{Span}
            (T)$ but $0 \notin \text{Span}(T+1)$.
    \end{itemize}

    \rule{\textwidth}{0.5pt}

    \textbf{2. Guidance from a Previous Thinking Model (Turn $t_{1} \leq t_{0}$):}

    Sometimes, you have access to a past output from a thinking model, computed
    at turn $t_{1}$ based on then-current observations. This guidance may no longer
    perfectly match the current situation but can still be valuable for decision-making.
    You can use this plan as a \textbf{strategic reference}, not a mandatory
    instruction. Consider how much of the original strategy is still valid under
    the current dynamics.

    \rule{\textwidth}{0.5pt}

    \textbf{3. Task (Turn $t_{0}$):}

    Choose \textbf{one} action $a_{t_0}\in \{U, D, S\}$ for the current turn,
    with the following considerations:

    \begin{itemize}
        \item \textbf{Collision Avoidance:} Ensure the action avoids both
            immediate and near-future collisions.

        \item \textbf{Strategic Consistency (Optional):} Refer to the thinking
            model's prior strategy. If the current environment still aligns with
            its assumptions, you may choose to continue along the same strategic
            direction. If not, adapt as needed.
    \end{itemize}

    \textbf{Answer Format}:

    \begin{center}
        \fbox{\begin{minipage}{0.3\textwidth}
            \centering
            $a_{t_0}$
        \end{minipage}}
    \end{center}

    \rule{\textwidth}{0.5pt}

    \textbf{4. Current State (Turn $t_{0}$):}

    \textbf{Current Turn:} $t_{0} = 10$

    \textbf{Player Position:} $(0, 6)$

    \textbf{Car State:}

    \begin{center}
        \scriptsize
        \begin{tabular}{|c|p{9cm}|}
            \hline
            \textbf{Freeway $k$} & \textbf{Cars (head $h$, tail $\tau$, direction $d$, speed $s$)}                              \\
            \hline
            1                    & $(48, 37, \text{right}, 12)$, $(0, -11, \text{right}, 12)$                                   \\
            \hline
            2                    & $(48, 1, \text{right}, 48)$                                                                  \\
            \hline
            3                    & $(-48, -25, \text{left}, 24)$                                                                \\
            \hline
            4                    & $(-48, -59, \text{right}, 12)$, $(-12, -23, \text{right}, 12)$, $(24, 13, \text{right}, 12)$ \\
            \hline
            5                    & $(-24, -35, \text{right}, 4)$                                                                \\
            \hline
            6                    & $(-24, -35, \text{right}, 4)$, $(52, 41, \text{right}, 4)$, $(16, 5, \text{right}, 4)$       \\
            \hline
            7                    & $(48, 37, \text{right}, 12)$, $(0, -11, \text{right}, 12)$                                   \\
            \hline
            8                    & $(48, 37, \text{right}, 12)$, $(0, -11, \text{right}, 12)$                                   \\
            \hline
        \end{tabular}
    \end{center}

    \textbf{Guidance from a Previous Thinking Model:} Turn $t_{1} = 10$

    Previous thinking process:
    \begin{guidancebox}
        \scriptsize

        First, I need to get the player from $(0,6)$ to $(0,9)$ in as few turns as
        possible, avoiding collisions with cars on the freeways $y=1$ to $y=8$. The
        current turn is $t_{1}=10$.

        The player can move up (U), down (D), or stay (S), changing $y$ by $+1$,
        $-1$, or $0$ respectively, and $y$ must stay between $0$ and $9$ inclusive.

        Cars move on specific freeways, and I need to avoid being on a freeway
        when a car is at position $x=0$ at that turn. The collision happens only
        if at turn $T$, the player is on freeway $y_{T}$, and for that freeway,
        there's a car where $0$ is in its span at that exact turn.

        Importantly, when I move to a new freeway, I'm not considered on that
        freeway until the next turn. So if I move from $y$ to $y+1$ at turn $T$,
        at turn $T$ I'm still on $y$, and I only occupy $y+1$ at turn $T+1$.
        That means I don't collide with cars on $y+1$ at turn $T$, only at turn $T
        +1$ if I'm still there.

        Now, the goal is to reach $y=9$. Since we start at $y=6$, and $y=9$ is
        three levels up, but I have to go through $y=7$ and $y=8$, and avoid the
        cars there.

        The minimal number of turns might be more than $3$ because I might need
        to wait or something to avoid collisions.

        $H$ is the number of turns, so from turn $t_{1}$ to $t_{1}+H-1$, so if I
        start at $t=10$, and I take $H$ turns, I end at turn $10+H-1$.

        I need $y_{10+H-1}= 9$.

        Since I start at $y=6$, and I need to get to $y=9$, which is $+3$, so minimum
        $H$ would be $3$ if I can go up three times without colliding, but I
        need to check if that's possible given the car positions.

        But the cars are moving, so I need to see when they are at $x=0$. First,
        let's understand the car movements.

        The car spans are given at turn t1=10.

        For each car, I need to find when 0 is in its span for future turns.

        The span changes linearly with time based on speed and direction.

        Let me denote the time elapsed from t1=10 as $\Delta t = T - 10$.

        For a car, at turn T, span is:

        If left-moving: [h - s*$\Delta t$, $\tau$ - s*$\Delta t$] since it moves
        left, decreasing x.

        In the rule: for left-moving, Span(T) = [h - s*(T-t1), $\tau$ - s*(T-t1)]

        Similarly for right-moving: Span(T) = [$\tau$ + s*(T-t1), h + s*(T-t1)]

        And collision if 0 is in Span(T) for the freeway we're on at turn T.

        Now, since the player is at y, and cars are per freeway, I need for each
        freeway k, and for each car on that freeway, find the times when 0 is in
        its span.

        Then, to avoid collision, when I'm on freeway k at turn T, none of the
        cars on k should have 0 in span at T.

        Now, for the path to y=9.

        Since y=9 is the top, and no cars above, but I need to pass through y=7
        and y=8.

        y=9 has no cars since cars are on y=1 to 8, so once I reach y=9, I'm safe.

        But I need to get there.

        Start at y=6 at t=10.

        I need to go to y=7, then y=8, then y=9.

        But I might stay or move down to avoid cars, but moving down might not
        help since I need to go up eventually.

        Since the grid is small, y from 0 to 9, and I start at 6, going to 9, it's
        only 3 steps up, so H min is 3, but if not possible, I need more turns by
        staying or moving down and up.

        ... ...
    \end{guidancebox}
\end{promptbox}
\newpage
\textbf{Prompt for Code-As-Policy}
\begin{promptbox}
    \footnotesize \textbf{Prompt:}

    Now a player is playing a multi-turn game, and suppose current turn is
    $\{t_{1}\}$. Given the initial position $(0, y_{t_1})$ on a 2D grid (vertical
    axis $y = 0, 1, \dots, 9$). You are tasked with implementing a Python
    function that determines the optimal next action to avoid collisions with
    moving cars and progress towards the goal position $(0, 9)$ in minimal turns.

    \rule{\textwidth}{0.5pt}

    \textbf{1. Game Dynamics:}

    \begin{itemize}
        \item \textbf{Player update:}\\ $y_{t+1}= y_{t} + \Delta y_{t}$, where
            \[
                \Delta y_{t}=
                \begin{cases}
                    +1 & \text{if }a_{t}= U \\
                    -1 & \text{if }a_{t}= D \\
                    0  & \text{if }a_{t}= S
                \end{cases}, \quad y_{t+1}\in [0,9]
            \]

        \item \textbf{Car update rules:}\\ For car $k$ on freeway $i$, suppose
            its head is at $h$, tail is at $\tau$ at turn $t_{1}$, and speed is
            $s$. Then at turn $T > t_{1}$, the car span becomes:
            \begin{itemize}
                \item Left-moving:
                    $\text{Span}(t_{1}) = [h, \tau] \rightarrow \text{Span}(T) =
                    [h - s (T-t_{1}), \tau - s (T-t_{1})]$

                \item Right-moving:
                    $\text{Span}(t_{1}) = [\tau, h] \rightarrow \text{Span}(T) =
                    [\tau + s (T-t_{1}), h + s (T-t_{1})]$
            \end{itemize}

        \item Collision occurs at turn $T$ only if $0 \in \text{Span}(T)$ for any
            car on freeway $y_{T}$.

        \item Note that if you decide to move to $y_{T+1}\neq y_{T}$ at turn $T$,
            you will \textbf{NOT} be considered to be on $y_{T+1}$ at turn $T$,
            thus will \textbf{NOT} be collided by cars on $y_{T+1}$ if $0 \in \text{Span}
            (T)$ but $0 \notin \text{Span}(T+1)$.
    \end{itemize}

    \rule{\textwidth}{0.5pt}

    \textbf{2. Task}

    You need to determine the best next action for the player by generating an \textbf{executable}
    Python function \texttt{next\_action(json\_state)} with the input \texttt{json\_state}
    representing the current game state as a JSON object. The function should analyze
    the game state and return the next action, represented as a single character
    string:
    \begin{itemize}
        \item 'U' for moving up (to $y + 1$)

        \item 'D' for moving down (to $y - 1$)

        \item 'S' for staying in the current position
    \end{itemize}

    Notice that the code will be executed in a loop, so it should return the next
    action each time it is called with the current game state, which will
    \textbf{change after each action}.

    \textbf{Input Format}

    \begin{lstlisting}[language=Python]
json_state = {
    'player_states': current_y_position,  # int: 0-9, 9 is the goal position
    'car_states': [  # list of tuples
        (lane, head, direction, speed, span),
        # lane: 1-8 (freeway number)
        # head: int, position of the car's head
        # direction: 'left' or 'right', tail = head + span if left, head - span if right
        # speed: int, speed of the car
        # span: int, span of the car, defined as the absolute difference between head and tail
    ],
    'turn': current_turn_number  # int: current turn
}
\end{lstlisting}

    \textbf{Output Format}

    Generate \textbf{two clearly labeled parts}:

    \textbf{Part 1: Summary}

    \begin{center}
        \fbox{\begin{minipage}{0.8\textwidth}
            One-sentence intent describing your strategy for the next actions
        \end{minipage}}
    \end{center}

    \textbf{Part 2: Python Function}

    \begin{lstlisting}[language=Python]
def next_action(json_state) -> str:
    """
    Returns one of the actions: 'U', 'D', or 'S' based on the current game state.

    Args:
        json_state: The current game state as JSON object

    Returns:
        str: Single character representing the next action ('U', 'D', 'S')
    """
    # Your logic here
    return action
\end{lstlisting}

    \textbf{Example Output}

    \textbf{Part 1: Summary}

    \begin{center}
        \fbox{\begin{minipage}{0.8\textwidth}
            Wait for the cars in lane 4 and 5 to pass, then move up to avoid
            collisions.
        \end{minipage}}
    \end{center}

    \textbf{Part 2: Python Function}

    \begin{lstlisting}[language=Python]
def next_action(json_state) -> str:
    # Implementation...
    return 'S'  # default action if no immediate threat
\end{lstlisting}

    \textbf{Current State}

    \begin{lstlisting}[language=Python]
{
    "player_states": 5,
    "car_states": [
        (1, -39, 'right', 3, 11), (2, 48, 'right', 12, 11),
        (2, -36, 'right', 12, 11), (2, 0, 'right', 12, 11),
        (3, 48, 'right', 12, 11), (3, -36, 'right', 12, 11),
        (3, 0, 'right', 12, 11), (4, 48, 'right', 12, 11),
        (4, -36, 'right', 12, 11), (4, 0, 'right', 12, 11),
        (5, -18, 'right', 6, 11), (6, -18, 'right', 6, 11),
        (7, -16, 'right', 4, 11), (8, -16, 'right', 4, 11)
    ],
    "turn": 8
}
\end{lstlisting}
\end{promptbox}

\section{Additional Experiment Results}
\label{sec:additional_results}

\subsection{Main Results}
\label{sec:main_results}

\begin{table}[H]
\centering
\caption{Complete agent performance across various cognitive load levels (Easy, Medium, Hard) with time pressure fixed at 8k tokens/step.}
\label{tab:complete_performance_data}

% Freeway Game
\textbf{Freeway}
\vspace{0.2cm}

\begin{tabular}{@{}lccccc@{}}
\toprule
\textbf{Cognitive Load} & \textbf{Code-as-Policy} & \textbf{Reactive (R1)} & \textbf{Reactive (V3)} & \textbf{Planning (R1)} & \textbf{\ourSystem} \\
\midrule
Easy   & 0.5393 & 0.2022 & 0.9775 & 0.3371 & 0.9551 \\
Medium & 0.6966 & 0.0000 & 0.3258 & 0.2247 & 0.8427 \\
Hard   & 0.5281 & 0.0000 & 0.0562 & 0.1011 & 0.5056 \\
\bottomrule
\end{tabular}

\vspace{0.3cm}
% Snake Game
\textbf{Snake}
\vspace{0.2cm}

\begin{tabular}{@{}lccccc@{}}
\toprule
\textbf{Cognitive Load} & \textbf{Code-as-Policy} & \textbf{Reactive (R1)} & \textbf{Reactive (V3)} & \textbf{Planning (R1)} & \textbf{\ourSystem} \\
\midrule
Easy   & 0.1719 & 0.1719 & 0.7694 & 0.0588 & 0.6931 \\
Medium & 0.1797 & 0.2188 & 0.4900 & 0.0256 & 0.5413 \\
Hard   & 0.0625 & 0.1406 & 0.2950 & 0.0137 & 0.3906 \\
\bottomrule
\end{tabular}

\vspace{0.3cm}
\textbf{Overcooked}
\vspace{0.2cm}

\begin{tabular}{@{}lccccc@{}}
\toprule
\textbf{Cognitive Load} & \textbf{Code-as-Policy} & \textbf{Reactive (R1)} & \textbf{Reactive (V3)} & \textbf{Planning (R1)} & \textbf{\ourSystem} \\
\midrule
Easy   & 0.4621 & 0.8193 & 0.9188 & 0.5379 & 1.0000 \\
Medium & 0.3724 & 0.6607 & 0.3664 & 0.0871 & 0.9152 \\
Hard   & 0.3661 & 0.2054 & 0.0877 & 0.0000 & 0.5982 \\
\bottomrule
\end{tabular}
\end{table}

\begin{table}[H]
\centering
\caption{Complete agent performance across time pressure levels (4k to 32k tokens/step) with cognitive load fixed at Medium.}
\label{tab:time_pressure}
\textbf{Freeway}
\vspace{0.2cm}

\begin{tabular}{@{}lccccc@{}}
\toprule
\textbf{Tokens/Turn} & \textbf{Code-as-Policy} & \textbf{Reactive (R1)} & \textbf{Reactive (V3)} & \textbf{Planning (R1)} & \textbf{\ourSystem} \\
\midrule
32k  & 0.9438 & 0.0000 & 0.2911 & 0.9621 & 0.9431 \\
16k  & 0.9551 & 0.0000 & 0.2911 & 0.9045 & 0.9347 \\
8k & 0.6966 & 0.0000 & 0.2911 & 0.2261 & 0.8469 \\
4k & 0.3483 & 0.0000 & 0.2911 & 0.1194 & 0.6166 \\
\bottomrule
\end{tabular}

\vspace{0.3cm}
\textbf{Snake}
\vspace{0.2cm}

\begin{tabular}{@{}lccccc@{}}
\toprule
\textbf{Tokens/Turn} & \textbf{Code-as-Policy} & \textbf{Reactive (R1)} & \textbf{Reactive (V3)} & \textbf{Planning (R1)} & \textbf{\ourSystem} \\
\midrule
32k  & 0.2109 & 0.0238 & 0.4844 & 0.9629 & 0.8281 \\
16k  & 0.2344 & 0.0238 & 0.4844 & 0.4043 & 0.7813 \\
8k & 0.1797 & 0.0238 & 0.4844 & 0.0254 & 0.5410 \\
4k & 0.0156 & 0.0238 & 0.4844 & 0.0176 & 0.4238 \\
\bottomrule
\end{tabular}

\vspace{0.3cm}
\textbf{Overcooked}
\vspace{0.2cm}

\begin{tabular}{@{}lccccc@{}}
\toprule
\textbf{Tokens/Turn} & \textbf{Code-as-Policy} & \textbf{Reactive (R1)} & \textbf{Reactive (V3)} & \textbf{Planning (R1)} & \textbf{\ourSystem} \\
\midrule
32k  & 0.5804 & 0.0000 & 0.3800 & 0.8371 & 0.9129 \\
16k  & 0.4621 & 0.0000 & 0.3800 & 1.0000 & 0.9375 \\
8k & 0.3661 & 0.0000 & 0.3800 & 0.0871 & 0.9152 \\
4k & 0.3724 & 0.0000 & 0.3800 & 0.0246 & 0.7087 \\
\bottomrule
\end{tabular}
\end{table}

\subsection{Significance Test}
\label{sec:significance_test}
We investigate the significance of advantage of \ourSystem\space over single-paradigm agents. We hypothesize that: (1) \ourSystem's advantage over reactive agents (V3) becomes more significant as cognitive load increases, and (2) its advantage over planning agents (R1) becomes more significant as time pressure increases.

To validate these hypotheses, we perform experiments across 3 cognitive load levels (Easy, Medium, Hard) and three 3 pressures (High: 32k tokens/step, Medium: 8k tokens/step, Low: 4k tokens/step). Below, we formally describe the validation procedure (1) and (2) can be tested similarly.

For each fixed cognitive load and time pressure condition, we analyze the paired score differences between \ourSystem and the reactive agent. Let $\mu_d$ denote the mean score difference between the paired observations (\ourSystem minus reactive agent). We pair observations that share the same environmental configuration seed, with each score calculated as the average across experimental runs within that configuration seed.
We formulate the hypotheses as follows:
\begin{itemize}
\item \textbf{Null Hypothesis}: $\mathbf{H_0}: \mu_d = 0$
\item \textbf{Alternative Hypothesis} $\mathbf{H_1}: \mu_d > 0$
\end{itemize}

The test statistic for the paired t-test is calculated as:
\[t = \frac{\bar{d}}{s_d/\sqrt{n}}\]

where $\bar{d}$ is the mean score difference, $s_d$ is the standard deviation of the differences, and $n$ is number of environment configuration seeds. We estimate the p-value based on the t-statistic with $n-1$ degrees of freedom. 

\begin{figure}[H]
    \centering
    \includegraphics[width=1\linewidth]{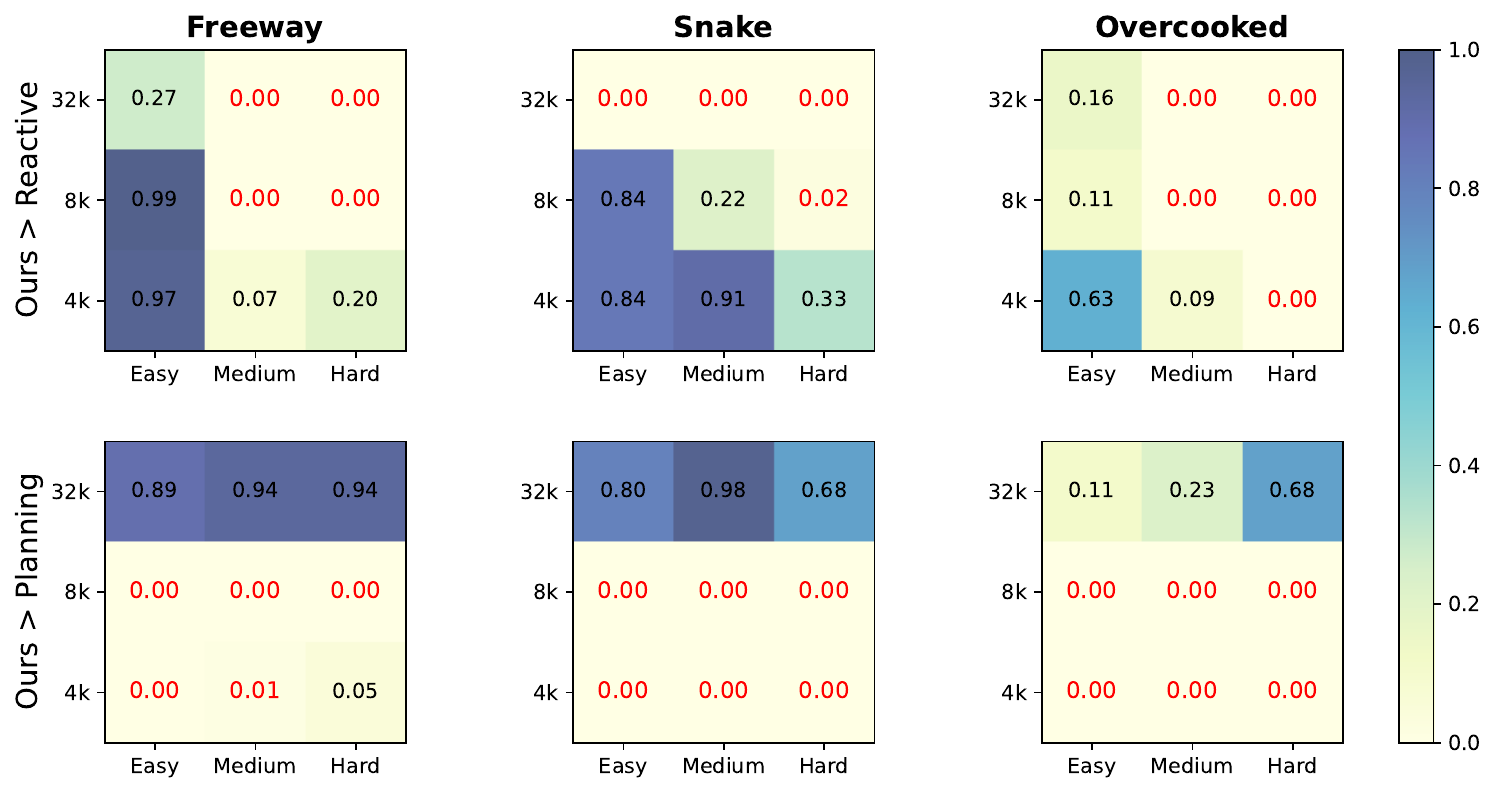}
    \caption{
    Statistical significance of \ourSystem's advantage over single-paradigm agents. Upper: Advantage over reactive agent (V3). Lower: Advantage over planning agent (R1). Numbers represent p-values under varying cognitive loads and time pressures, with red indicating statistical significance ($p < 0.05$). The advantage of \ourSystem generally increases with both cognitive load and time pressure.}
    \label{fig:significance_test}
%    \vspace{-20pt}
\end{figure}

Figure~\ref{fig:significance_test} presents the p-values across different conditions, with statistical significance assessed at $\alpha = 0.05$. The results show that \ourSystem's advantage generally becomes statistically significant as cognitive load and time pressure increase.

\subsection{Results of Other Models}
\label{sec:gemini}

Since \ourSystem relies on transparent reasoning trajectories, which are only available in open-source models, our primary experiments are conducted using the state-of-the-art open-source DeepSeek models. Besides using DeepSeek-V3 and DeepSeek-R1 as the reactive and planning model, we also conduct experiments with DeepSeek-V3.2, where non-thinking mode is used as reactive and thinking mode is used as planning. The results are shown in Tab.~\ref{tab:v32-results} and Tab.~\ref{tab:v32-walltime-results}.

\begin{table}
\centering
\caption{DeepSeek-V3.2 based agents' performance across different time pressure levels (4k to 32k tokens/step) with cognitive load fixed at High.}
\label{tab:v32-results}
\textbf{Freeway}
\vspace{0.2cm}

%freeway {'Reactive (Non-Thinking)': [0.4659550561797753, 0.4659550561797753, 0.4659550561797753, 0.4659550561797753], 'Planning (Thinking)': [0.8539325842696629, 0.46494382022471914, 0.10247191011235954, 0.0], 'AgileThinker': [0.8889887640449439, 0.832808988764045, 0.6351685393258427, 0.44662921348314605]}
%snake {'Reactive (Non-Thinking)': [0.255625, 0.255625, 0.255625, 0.255625], 'Planning (Thinking)': [0.738125, 0.390625, 0.055, 0.0075], 'AgileThinker': [0.57, 0.5349999999999999, 0.345625, 0.27375]}
%overcooked {'Reactive (Non-Thinking)': [0.5246428571428571, 0.5246428571428571, 0.5246428571428571, 0.5246428571428571], 'Planning (Thinking)': [0.8371428571428572, 0.45535714285714285, 0.013392857142857142, 0.0067857142857142855], 'AgileThinker': [0.8550000000000001, 0.84375, 0.65625, 0.44857142857142857]}
\begin{tabular}{@{}lccccc@{}}
\toprule
\textbf{Tokens/Turn}  & \textbf{Reactive (thinking off)} & \textbf{Planning (thinking on)} & \textbf{\ourSystem} \\
\midrule
32k & 0.4659 & 0.8539 & 0.8889 \\
16k & 0.4659 & 0.4649 & 0.8328 \\
8k & 0.4659 & 0.1025 & 0.6352 \\
4k & 0.4659 & 0.0000 & 0.4466 \\
\bottomrule
\end{tabular}

\vspace{0.4cm}
\textbf{Snake}
\vspace{0.2cm}

\begin{tabular}{@{}lccccc@{}}
\toprule
\textbf{Tokens/Turn} & \textbf{Reactive (thinking off)} & \textbf{Planning (thinking on)} & \textbf{\ourSystem} \\
\midrule
32k & 0.2556 & 0.7381 & 0.5700 \\
16k & 0.2556 & 0.3906 & 0.5350 \\
8k & 0.2556 & 0.0550 & 0.3456 \\
4k & 0.2556 & 0.0075 & 0.2738 \\
\bottomrule
\end{tabular}

\vspace{0.4cm}
\textbf{Overcooked}
\vspace{0.2cm}

\begin{tabular}{@{}lccccc@{}}
\toprule
\textbf{Tokens/Turn}  & \textbf{Reactive (thinking off)} & \textbf{Planning (thinking on)} & \textbf{\ourSystem} \\
\midrule
32k & 0.5246 & 0.8371 & 0.8550 \\
16k & 0.5246 & 0.4554 & 0.8438 \\
8k & 0.5246 & 0.0134 & 0.6563 \\
4k & 0.5246 & 0.0068 & 0.4486 \\
\bottomrule
\end{tabular}
\end{table}

\begin{table}
\centering
   \caption{DeepSeek-V3.2 based agents' performance with time pressure of 350 seconds per step and high cognitive load (Hard difficulty), confirming \ourSystem's advantage persists in wall-time constraints.}
    \label{tab:v32-walltime-results}

   \begin{tabular}{llll}
        \toprule \textbf{Environment} & \textbf{Reactive (thinking off)} & \textbf{Planning (thinking on)} & \textbf{\ourSystem} \\
        \midrule 
        Freeway              & 0.3665        & 0.0        & 0.7022        \\
        Snake                         & 0.3050        & 0.0863        & 0.6013    \\
        Overcooked                    & 0.5246        & 0.0        & 0.6675        \\
        \bottomrule
    \end{tabular}
\end{table}

To assess the generalizability of our approach, we also evaluate reactive and planning agents using other models, such as Gemini-2.5-Flash, which features an intrinsic budget control function. For reactive agents, we disable extended thinking, while for planning agents, thinking is enabled.

However, we cannot directly implement \ourSystem with Gemini due to the lack of access to its reasoning traces. Instead, we approximate the system using a reduced design: the reactive agent references the \emph{final} output of the planning agent after it completes reasoning, rather than accessing \emph{partial} reasoning traces during streaming. Despite this limitation, the results in Tab.~\ref{tab:gemini-performance} consistently show that combining reactive and planning paradigms improves performance across all levels of time pressure.

\begin{figure}[H]
    \centering
    \includegraphics[width=1\linewidth]{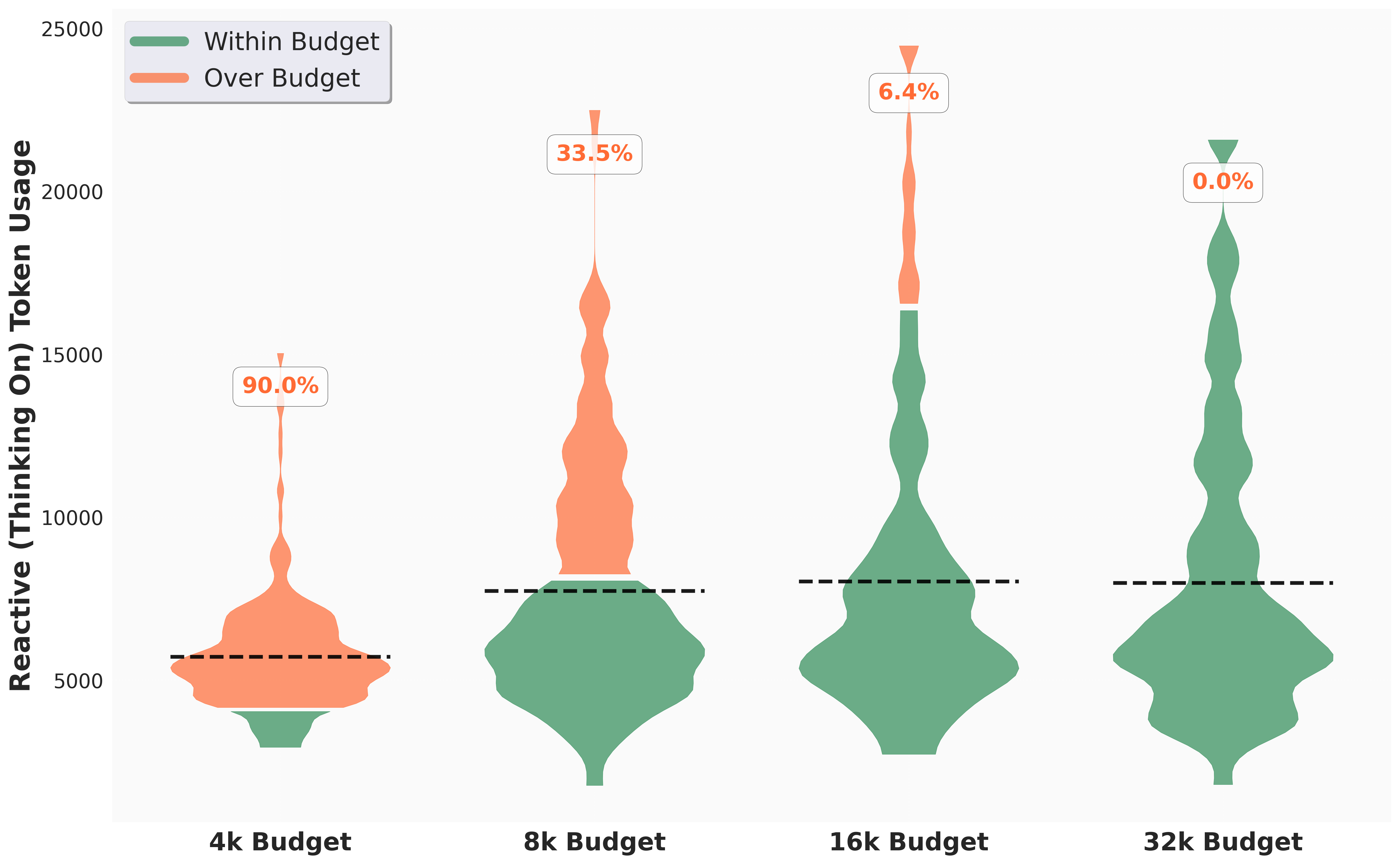}
    \caption{
    To test the effectiveness of thinking budget control, we set the thinking budget of Gemini-2.5-Flash at 4k, 8k, 16k and 32k to investigate the distribution of response length.  
    The distribution shows that even Gemini-2.5-Flash is aware of the budget, it cannot precisely regulate response token count, often generating excessive tokens. }
    \label{fig:token_usage_vs_budget}
\end{figure}

Additionally, we observe that Gemini-2.5-Flash's built-in budget control struggles to precisely regulate token usage, often exceeding the allocated budget (Figure~\ref{fig:token_usage_vs_budget}). This highlights the ongoing challenge that LLMs face in controlling computational costs during real-time reasoning.
%Specifically, we turn thinking off in the reactive paradigm and turn thinking on in the planning paradigm.
%Meanwhile, we must approximate \ourSystem with a weaker design due to the lack of accessible reasoning traces. Therefore, we cannot assume the response is available in a streaming way, but available only after it's complete.
%The results in Table~\ref{tab:gemini-performance} show similar trends: combining reactive and planning paradigms consistently improves performance under all levels of time pressure.
%We also find that the budget control method provided by Gemini-2.5-Flash cannot precisely regulate token usage and often exceed the allocated budget (Figure~\ref{fig:token_usage_vs_budget}). This further suggests that current LLMs struggle to account for time cost.

\begin{table}[H]
\centering
\caption{\textbf{Performance of Gemini-2.5-Flash (Medium-difficulty Freeway) under various agent designs.} 
We implement budget control by setting the thinking budget equal to the time pressure budgets. 
Since Gemini-2.5-Flash's internal reasoning traces are not public, we cannot apply \ourSystem\ directly. 
Instead, we reduce Reactive + Planning to allowing reactive thread to reference the \emph{final} non-thinking tokens produced by planning thread after its reasoning is completed.}
\label{tab:gemini-performance}

\begin{tabular}{@{}lccccc@{}}
\toprule
\makecell{\textbf{Tokens/Step}} & \makecell{\textbf{Reactive}\\\textbf{(thinking off)}} & \makecell{\textbf{Reactive}\\\textbf{(thinking on + budget control)}} & \makecell{\textbf{Planning}\\\textbf{(thinking on)}} & \makecell{\textbf{Reactive+}\\ \textbf{ Planning}} \\

\midrule
32k   & 0.12 & 0.93 & 0.93 & 0.92\\
16k & 0.12 & 0.76 & 0.70 & 0.70\\
8k   & 0.12 & 0.09 & 0.25 & 0.31 \\
4k   & 0.12 & 0.00 & 0.05 & 0.26 \\
\bottomrule
\end{tabular}
\end{table}

\subsection{Code-as-Policy}
\label{sec:code_as_policy}
We analyze generated code to understand the limitations of code-as-policy
approaches. Below are representative examples from each environment. All code is
directly generated by DeepSeek-R1 with some helper functions omitted for brevity.

\subsubsection*{Freeway}
\begin{python}
    def next_action(json_state) -> str: # Extract game state # Helper functions
    ... t0, y0, cars = extract_game_state(json_state) max_time = t0 + 50 # Reasonable
    time horizon visited = set() queue = deque() start = (t0, y0, []) # (current_turn,
    y, path_actions) visited.add((t0, y0)) queue.append(start)

    while queue: # BFS path finding ... # BFS found no path: choose safest
    immediate next action ...
\end{python}
\textbf{Freeway} has a simple, efficient solution: breadth-first search (BFS) to
explore all possible movement sequences across lanes, and the state space is
very small (time $\times$ player position). Therefore, LLMs can implement naive
search strategies to get optimal performance.

\subsubsection*{Snake}
\begin{python}
    def next_action(json_state) -> str: # Extract game state # Helper functions
    ... # Main logic: choose best move for current state safe_moves = generate_safe_moves(json_state)
    if not safe_moves: return 'U' # no safe move

    best_move = None best_score = -float('inf') depth = 5 # Limited depth search

    for move in safe_moves: next_state, immediate_reward = simulate_move(json_state,
    move) future_reward = search(next_state, depth - 1) total_reward = immediate_reward
    + future_reward if total_reward > best_score: best_score = total_reward
    best_move = move

    return best_move
\end{python}

\textbf{Snake} does not have an algorithmic solution with bounded complexity, forcing
LLMs to resort to heuristic approaches. The generated code typically implements
limited-depth search (5 turns in this example) to evaluate immediate rewards, but
this constrained horizon proves insufficient for complex scenarios. Long-term
consequences - such as distant food opportunities or collision risks that emerge
beyond the search depth - remain invisible to the policy.

\subsubsection*{Overcooked}
\begin{python}
    def next_action(json_state: dict) -> str: # Extract game state # Helper functions
    ... alice_held_obj = alice.get('held_object')

    # Decision logic if alice_held_obj and alice_held_obj.get('name') == 'onion':
    # Find available pots and add ingredients ...

    elif alice_held_obj and alice_held_obj.get('name') == 'plate': # Pick up
    ready soup ... elif not alice_held_obj: # Priority 1: Get plate for ready
    soups ... # Priority 2: Get ingredients for cooking ...

    return 'S' # Stay as default
\end{python}

\begin{python}
    def next_action(json_state: dict) -> str: if alice_held_obj.get('name') == 'onion':
    pot = find_pot() add_ingredient(pot) elif alice_held_obj.get('name') == 'plate':
    soup = find_ready_soup() pickup_ready_soup() elif not alice_held_obj: # Priority
    1: Get plate for ready soups # Priority 2: Get ingredients for cooking ... return
    'S' # Stay as default
\end{python}

\textbf{Overcooked} has complex game context, demanding that players infer
partner intentions through action history and plan coordinated responses based
on current state. However, the generated code considers only limited context (e.g.,
what Alice is currently holding). This narrow focus leads to suboptimal decisions.
For instance, when the agent holds an onion but all pots are occupied (perhaps
by Bob), it simply stays idle rather than dropping the onion to pick up a plate
and serve ready soup.

\begin{python}
    def next_action(json_state: dict) -> str: # Extract game state # Helper functions
    ...

    pots = [tuple(pos) for pos in layout['Pot']] ready_soups = [obj for obj in objects
    if obj.get('name') == 'soup' and obj.get('is_ready', False)] cooking_soups =
    [obj for obj in objects if obj.get('name') == 'soup' and obj.get('is_cooking',
    False)]

    alice_held_obj = alice.get('held_object')

    # Priority 1: Serve ready soup if ready_soups: if alice_held_obj and alice_held_obj.get('name')
    == 'soup': # Deliver soup to serving counter ... elif alice_held_obj and
    alice_held_obj.get('name') == 'plate': # Pick up ready soup from pot ... else:
    # Get plate from dispenser ...

    # Priority 2: Handle cooking soups elif cooking_soups: soup = cooking_soups[0]
    if soup['remaining_cooking_tick'] <= 1 and not alice_held_obj: # Get plate for
    soon-to-be-ready soup ... # Priority 3: Start cooking (add ingredients to empty
    pot) idle_pot = next((pot for pot in pots if not any(obj.get('position') == list(pot)
    for obj in objects)), None) if idle_pot: if held_obj and held_obj.get('name')
    == 'onion': # Add ingredient to pot ... else: # Get ingredient from dispenser
    ...

    return 'S' # Stay as default
\end{python}

Beyond ignoring broad game context, the generated code exhibits inconsistent
goal prioritization that leads to oscillatory behavior. Consider a scenario where
Alice holds an onion and ready soup is available: the first code prioritizes
adding the onion to an empty pot, while the second prioritizes serving the ready
soup. These conflicting objectives cause the agent to alternate between
incompatible actions. This demonstrates a fundamental limitation of code-as-policy
approaches: while they can encode local heuristics effectively, they struggle to
maintain coherent long-term strategies in environments requiring dynamic coordination
and context-aware decision making.

\subsection{Performance under Limited Throughput}
\begin{table}[H]
    \begin{center}
        \begin{tabular}{lllll}
            \toprule \bf Game & \bf Parallel Threads & \bf Concurrent Threads & \bf Reactive (V3) & \bf Planning (R1) \\
            \midrule Freeway  & 0.84           & 0.72                   & 0.32              & 0.22              \\
            Snake             & 0.54           & 0.58                   & 0.49              & 0.03              \\
            Overcooked        & 0.92           & 0.85                   & 0.37              & 0.09              \\
            \bottomrule
        \end{tabular}
    \end{center}
    \caption{Performance of \ourSystem\space with abundant (parallel threads) and limited (concurrent threads) throughput.}
    \label{tab:parallel_concurrent}
\end{table}

%\begin{figure}[H]
%    \centering
%    \includegraphics[width=1\linewidth]{figures/concurrent.pdf}
%    \caption{Performance of agent systems with parallel and concurrent execution. }
%    \label{fig:parallel_concurrent}
%\end{figure}

While our main experiments assume parallel execution with independent throughput
for each subsystem, practical deployments may face resource constraints. In this
section, we investigate whether \ourSystem remains effective when LLM and LRM
share computational resources through concurrent switching rather than
parallelism. Specifically, we evaluate the performance of agent systems under
equivalent token throughput, implementing the \ourSystem as alternating inference between reactive and planning thread (concurrent generation) rather than simultaneous generation.

Figure~\ref{tab:parallel_concurrent} compares concurrent \ourSystem\space against parallel \ourSystem\space and single-model baselines. Even with equivalent throughput, concurrent \ourSystem\space significantly outperform
both reactive agents and planning agents. While parallel execution offers modest performance improvements over concurrent execution, the gains are relatively small compared to the fundamental advantage of \ourSystem. This indicates
that the primary benefit stems from cognitive specialization rather than
computational resources, and \ourSystem\space remain effective for resource-constrained
deployments.

\section{Walltime Experiments}
\begin{figure}[H]
    \centering
    \includegraphics[width=0.5\textwidth]{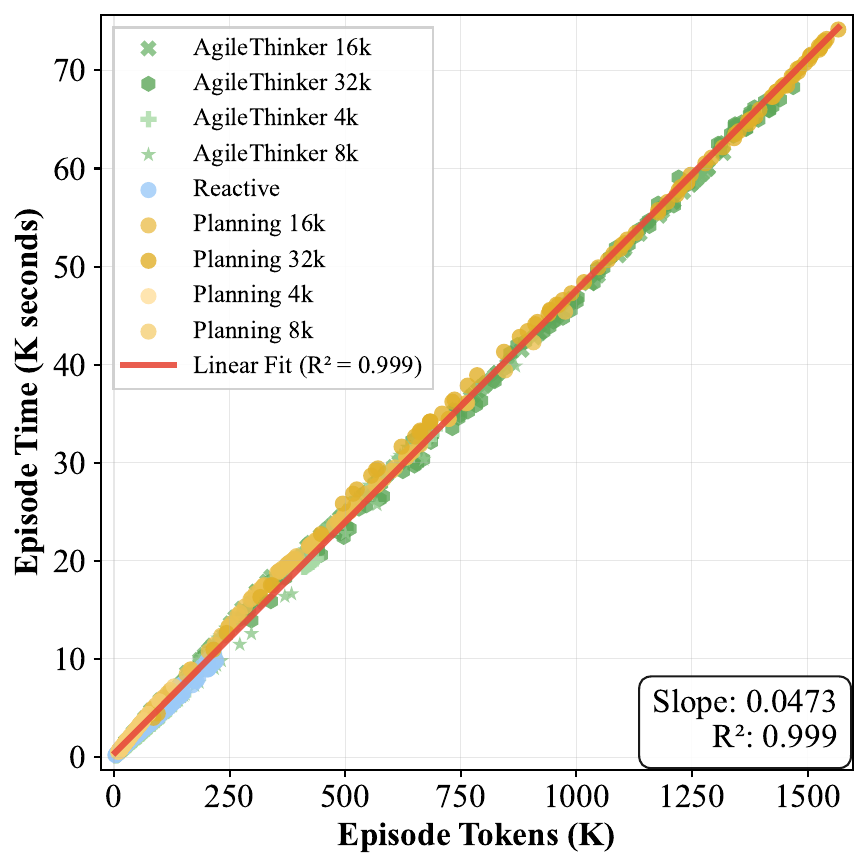}
    \caption{Almost linear correlation between generated token count and wall-clock time
    using DeepSeek official API, demonstrating the validity of our token-as-time
    abstraction. Here the numbers after agent  methods, e.g. 4k, 8k, refer to the corresponding environment time pressure budgets.}
    \label{fig:wall_time_align}
\end{figure}
\end{document}